\documentclass{article}
\pdfoutput=1
\newcommand{\method}{\textbf{Neuron Steadiness Regularization}}
\newcommand{\normethod}{neuron steadiness regularization }

\usepackage{floatrow}
\newfloatcommand{capbtabbox}{table}[][\FBwidth]
\usepackage[preprint,nonatbib]{neurips_2022}
\usepackage{wrapfig,lipsum,booktabs}

\usepackage[utf8]{inputenc} 
\usepackage[T1]{fontenc}    
\usepackage{hyperref}       
\usepackage{url}            
\usepackage{booktabs}       
\usepackage{nicefrac}       
\usepackage{microtype}      
\usepackage{xcolor}         

\usepackage{algorithm}
\usepackage{algorithmic}
\usepackage{amsthm}

\usepackage{enumitem}
\usepackage{makecell}
\usepackage{xspace}
\usepackage{amsmath,amssymb,amsfonts}
\usepackage{graphicx}
\usepackage{textcomp}
\usepackage{subfigure}
\usepackage{tcolorbox}
\usepackage{booktabs}
\usepackage{tabularx}
\usepackage{lipsum}
\usepackage{multirow}
\usepackage{color, xcolor}
\usepackage{balance}

\theoremstyle{plain}
\newtheorem{theorem}{Theorem}[section]

\newtheorem{lem}[theorem]{Lemma}

\theoremstyle{definition}

\theoremstyle{remark}

\title{Neuron with Steady Response Leads to Better Generalization}

\author{%
  Qiang Fu \thanks{These authors contributed equally to the work.} 
  ~ \thanks{Corresponding Author}\\
  Microsoft Research Asia\\
  Beijing, China\\
  \texttt{qifu@microsoft.com} \\
  \And
  Lun Du \footnotemark[1]\\
  Microsoft Research Asia\\
  Beijing, China\\
  \texttt{lun.du@microsoft.com} \\
  \AND
  Haitao Mao \footnotemark[1]~~\thanks{Work performed during the internship at MSRA.}\\
  Michigan State University \\
  Michigan, U.S. \\
  \texttt{haitaoma@msu.edu} \\
  \And
  Xu Chen \footnotemark[1]\\
  Microsoft Research Asia\\
  Beijing, China\\
  \texttt{xu.chen@microsoft.com} \\
  \And
  Wei Fang \footnotemark[1]~ \footnotemark[3]\\
  Tsinghua University\\
  Beijing, China\\
  \texttt{fw17@mails.tsinghua.edu.cn} \\
  \And
  Shi Han \\
  Microsoft Research Asia\\
  Beijing, China\\
  \texttt{shihan@microsoft.com} \\
  \And
  Dongmei Zhang \\
  Microsoft Research Asia\\
  Beijing, China\\
  \texttt{dongmeiz@microsoft.com} \\
}

\begin{document}
\maketitle
\begin{abstract}
Regularization can mitigate the generalization gap between training and inference by introducing inductive bias. 
Existing works have already proposed various inductive biases from diverse perspectives. 
However, none of them explores inductive bias from the perspective of class-dependent response distribution of individual neurons. 
In this paper, we conduct a substantial analysis of the characteristics of such distribution. 
Based on the analysis results, we articulate the Neuron Steadiness Hypothesis: the neuron with similar responses to instances of the same class leads to better generalization. 
Accordingly, we propose a new regularization method called Neuron Steadiness Regularization (NSR) to reduce neuron intra-class response variance. Based on the Complexity Measure, we theoretically guarantee the effectiveness of NSR for improving generalization.
We conduct extensive experiments on Multilayer Perceptron, Convolutional Neural Networks, and Graph Neural Networks with popular benchmark datasets of diverse domains, which show that our Neuron Steadiness Regularization consistently outperforms the vanilla version of models with significant gain and low additional computational overhead. 
\end{abstract}

\section{Introduction}
\label{sec:introduction}
Deep Neural Network (DNN) achieves state-of-the-art results in a wide range of areas and has various applications across industries, including self driving cars \cite{rao2018deep}, virtual assistants \cite{rawassizadeh2019manifestation}, intelligent healthcare \cite{miotto2018deep}, personalized recommendation \cite{naumov2019deep}, etc. DNN's success usually relies on a plenty amount of training data. However, DNN's generalization is often hampered in many domains where training data is insufficient because data annotation is labor-intensive and expensive. 

Regularization is a popular method that helps to improve generalization through introducing inductive bias. Regularization is one of the key elements of machine learning, particularly of deep learning \cite{goodfellow2016deep}. Specifically, inductive bias represents assumptions about the model properties other than the consistency of outputs with targets. There have been tremendous efforts in identifying such desired properties, which results in a series of widely used regularization methods. For example, L2 Regularization \cite{plaut1986experiments, lang1990dimensionality} penalizes large norms of model weights, which puts constraints on ``parameter scale". L1 regularization improves ``sparseness" by rewarding zero weight or neuron response. Jacobian regularization \cite{sokolic2017robust, hoffman2019robust} minimizes the norm of the input-output Jacobian matrix to improve the ``smoothness" of the learned mapping function. Orthogonal regularization \cite{cui2020towards, brock2016neural} enlarges ``weight diversity" to reduce the feature redundancy. Batch Normalization \cite{ioffe2015batch} promotes ``training dynamics stability" by reducing the internal covariate shift.

Although the existing works have already proposed various inductive biases from diverse perspectives, including the aforementioned ``parameter scale", ``sparseness", ``smoothness", ``weight diversity" and ``training dynamics stability", to the best of our knowledge, there is no work to explore inductive bias from the perspective regarding the characteristics of neuron response distribution on each class. From another point of view, the existing works leverage the information related to weights (parameter scale regularization), weight correlations (orthogonal regularization), derivatives of mapping function (smoothness regularization), collective neurons responses (sparseness regularization, Batch Normalization), but none of them considers the intra-class response distribution of individual neurons.   

\begin{figure}[!ht]
    \centering 
    \includegraphics[width=1\textwidth]{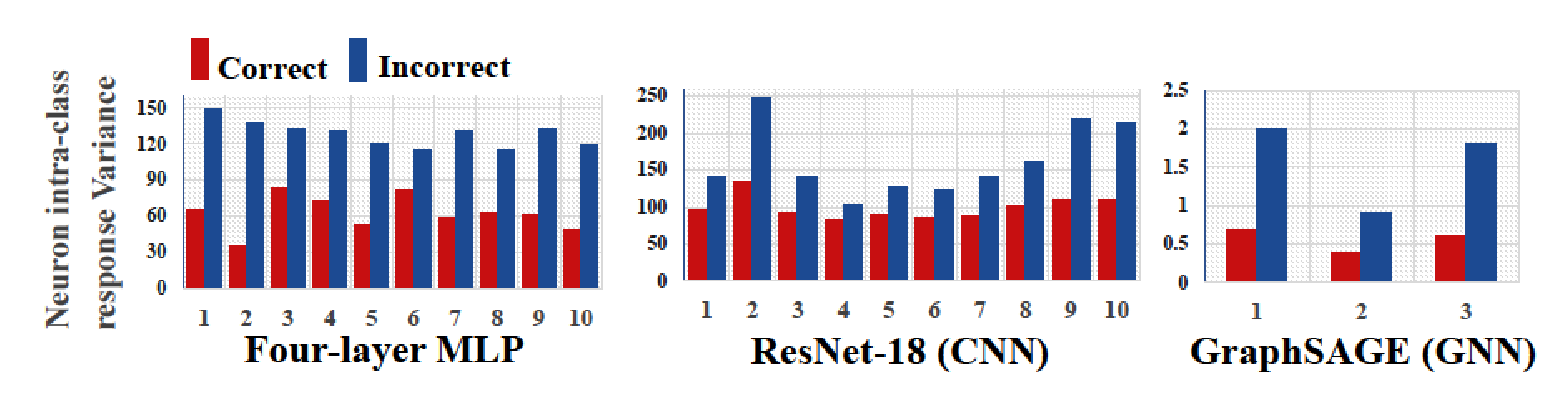}
    \caption{\label{fig:correctness}Comparison of the intra-class response variance of correctly and incorrectly classified testing samples for different architectures: four-layer MLP for MNIST, ResNet-18 for CIFAR-10, and GraphSAGE for PubMed. The horizontal axis and the vertical axis represent class indexes and the value of intra-class response variance, respectively. Each bar represents the intra-class response variance aggregated from all neurons in the penultimate layer.}
    
\end{figure}

\begin{figure}[!ht]

    \centering
    \subfigure[Testing accuracy]{
    \begin{minipage}[b]{0.31\textwidth}
    \centering
    \includegraphics[width=1\textwidth]{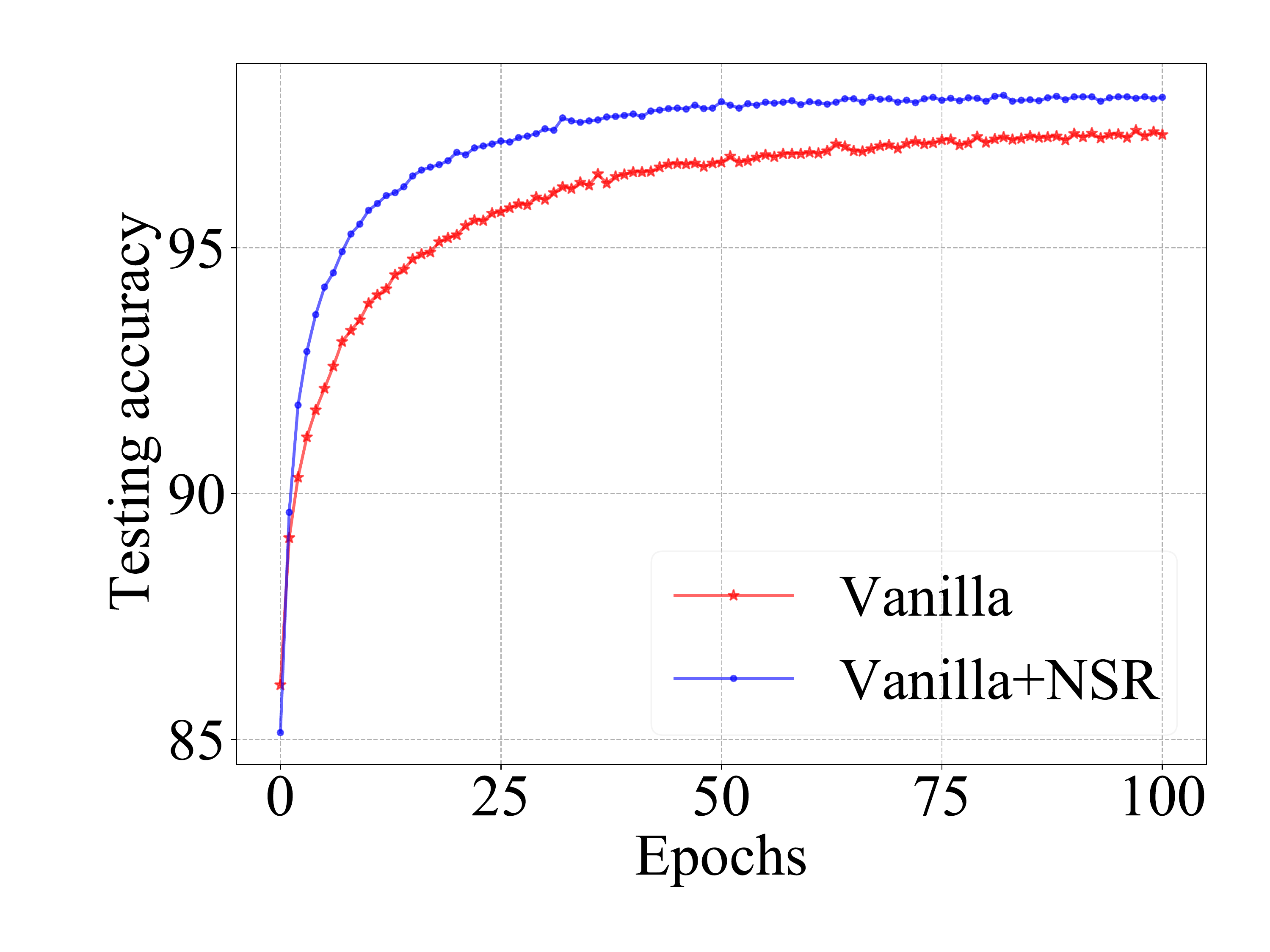}
    \end{minipage}
    }
    \subfigure[Cross entropy loss]{
        \begin{minipage}[b]{0.31\textwidth}
        \centering
        \includegraphics[width=1\textwidth]{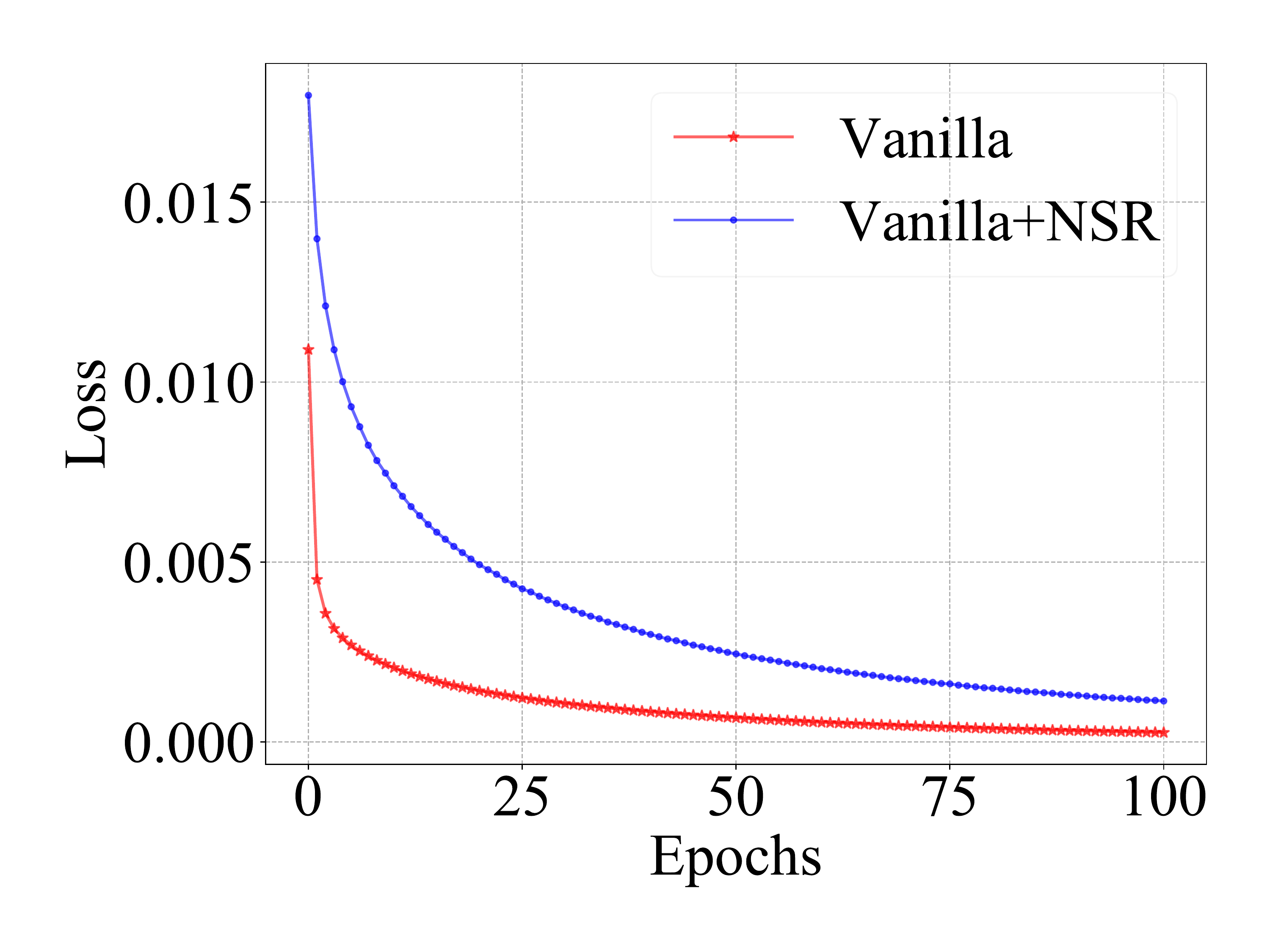}
        \end{minipage}
    }
    \subfigure[Average intra-class variance]{
        \begin{minipage}[b]{0.31\textwidth}
        \centering
        \includegraphics[width=1\textwidth]{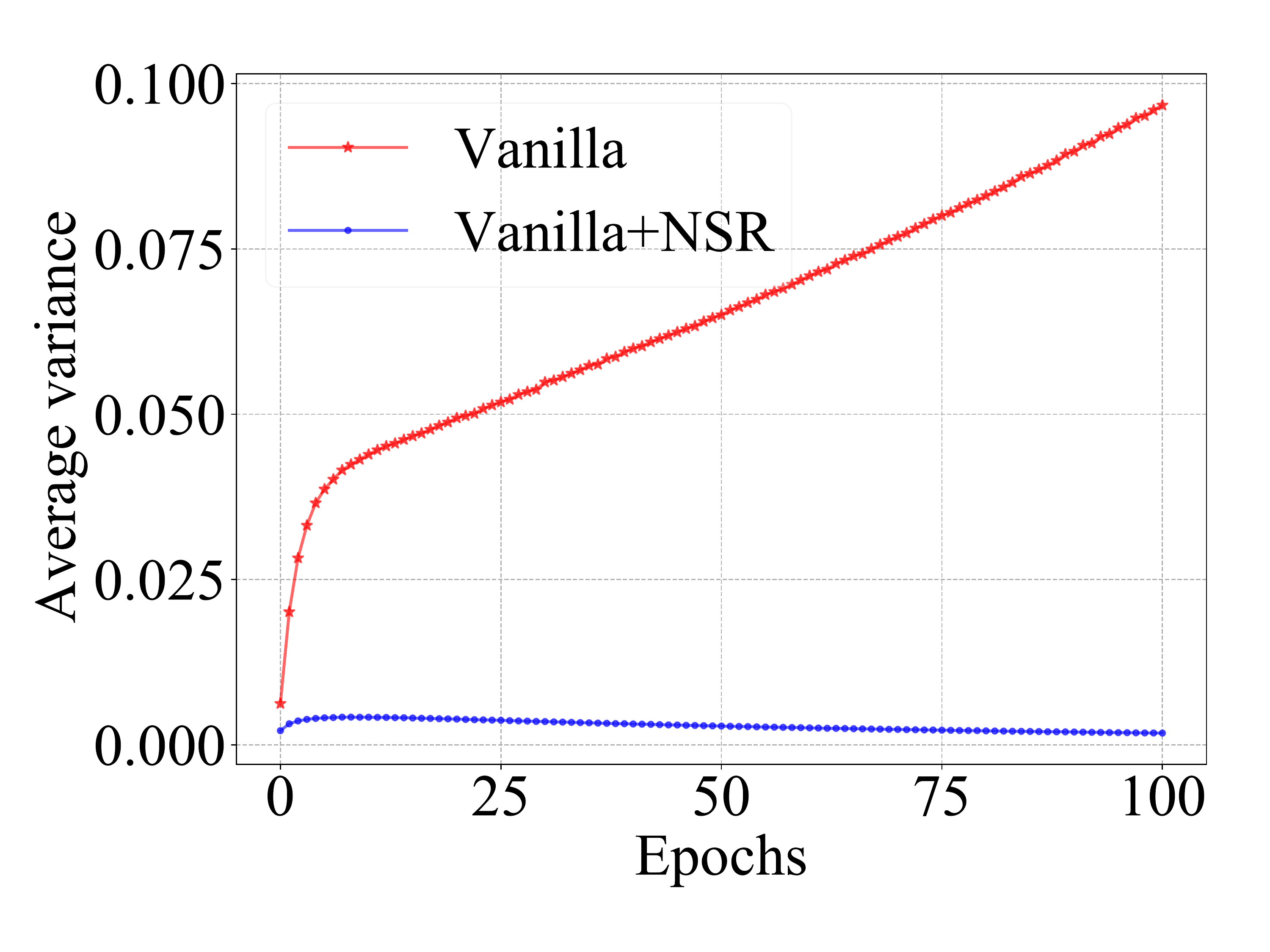}
        \end{minipage}
    }
    
    \caption{\label{fig:hyperparam}Training procedure of the vanilla four-layer MLP and the four-layer MLP with NSR on MNIST. (a) represents the testing accuracy; (b) and (c) illustrate the corresponding cross entropy loss and average intra-class response variance of these two models on the training set.}
\end{figure}

In this paper, we study the characteristics of the intra-class response distribution of each individual neuron to identify the new regularization method. In more detail, for each individual neuron, we analyze the variance of its response to samples of the same class, which is called \textbf{neuron intra-class response variance}. We find that such intra-class response variance has an obvious correlation with classification correctness. As shown in Figure \ref{fig:correctness}, we find the correctly classified samples usually have smaller intra-class response variance compared to the misclassified samples. Besides, it can be observed that the vanilla model with cross entropy as the optimization target usually could not control intra-class response variance well, as shown in Figure \ref{fig:hyperparam}, which leaves a potential improvement space for the regularization. The details of experiments and observations are explained in Section \ref{sec:Observations}.

Based on these observations, we articulate a \textbf{Neuron Steadiness Hypothesis}: neuron with similar responses to instances of the same class, i.e., smaller neuron intra-class response variance can lead to better generalization. 
Accordingly, we propose the regularization method called \textbf{Neuron Steadiness Regularization (NSR)} to improve generalization by penalizing large neuron intra-class response variance.

Based on the Complexity Measure theory in Section \ref{sec:theoretical_analysis}, we conduct theoretical analysis of the effectiveness of NSR for improving generalization. In addition, our regularization method shows significant improvements in various network architectures, including Graph Neural Networks (GNN), Convolution Neural Networks (CNN), and Multilayer Perceptron (MLP). 

To sum up, our contributions are as follows:
\begin{itemize}[leftmargin=*]
    \item We articulate a Neuron Steadiness Hypothesis and demonstrate its validity. It provides a new regularization perspective based on the neuron-level class-dependent response distribution. 
    \item We propose a new regularization method, Neuron Steadiness Regularization, to improve generalization ability. The method is computationally efficient and general enough to be applied to various architectures and tasks. Theoretical analyses guarantee its effectiveness.
    \item Extensive experiments are conducted on multiple types of datasets like images, citation graphs and product graphs, with various network architectures including GNN, CNN and MLP. Significant  improvements evidently verify the effectiveness of Neuron Steadiness Regularization. 
\end{itemize}

\section{Observations}\label{sec:Observations}
In this section, we verify the Neuron Steadiness Hypothesis by experiments with the following identified observations. 

\subsection{\textbf{Correlation between neuron intra-class response variance and classification correctness}}

The neuron intra-class response variance is derived from neuron response distribution 
which could be obtained by recording neuron responses when we feed input samples to the model. In this paper, for the neuron with ReLU as its activation function, we do not take its zero-response into account for calculation, because zero-response corresponds to the inactivated state where the neuron does not respond at all. In other words, for neurons with ReLU as the activation function, we only record its non-zero responses to represent its response distribution. 

With the obtained response distribution of any given neuron, it is relatively straightforward to calculate our discussed statistics, i.e., intra-class response variance, which is 
the mean squared deviation of the neuron response from the mean of the intra-class response. 
Then, for each neuron, we calculate intra-class response variance corresponding to the correctly classified samples and misclassified samples, respectively. Finally, we aggregate such two respective variances of all neurons in the penultimate layer separately and present them in Figure \ref{fig:correctness}.  

Figure \ref{fig:correctness} shows that for different networks, the average intra-class response variance of correctly classified samples is smaller than that of misclassified ones on arbitrary class. It 
indicates the strong correlation between classification correctness and neuron intra-class response variance. 

\subsection{\textbf{Dynamics of neuron intra-class response variance during training procedure}} \label{sec:2.2}

We investigate the tendency of neuron intra-class response variance along with the training procedure. For comparison, we perform the analysis on the vanilla model and the model with our proposed neuron steadiness regularization. We calculate the intra-class response variance of each neuron on the entire training set after each training epoch. Then, the intra-class response variance of all neurons is averaged and denoted as average variance in Figure \ref{fig:hyperparam} (c). We also show the testing accuracy and the training cross entropy loss in Figure \ref{fig:hyperparam} (a) and (b), respectively. Other architectures demonstrate similar tendencies and can be found in Appendix \ref{app:Dynamics}.

From Figure \ref{fig:hyperparam}, we could see that the cross entropy objective keeps being optimized during the training procedure which leads to increasing classification accuracy for both models. However, for the vanilla version, the average neuron intra-class response variance is growing larger because the model training does not impose constraints on neuron intra-class response variance. 
For the model trained with our proposed regularization, the neuron intra-class response variance is controlled and decreases after a few epochs. More importantly, the learned model with well-controlled intra-class response variance has higher testing accuracy than the vanilla version although its corresponding cross entropy loss is even larger. 
In addition, we could also see that regulating neuron intra-class response variance may also help the optimization procedure to achieve higher accuracy in earlier epochs. 

To conclude all the above observations, it is reasonable to design the regularization based on the neuron steadiness hypothesis, i.e., reducing neuron intra-class response variance, for better generalization.
\section{Method}
In this section, we first describe the proposed \method. Then we further introduce several techniques for computational efficiency.
\subsection{Definition of \method}
\textbf{\method} (NSR) for a specific neuron is defined as the summation of intra-class response variances of different classes. The NSR for the $n$-th neuron can be formulated as:
\begin{equation}
\label{equ:neuron_stable_reg}
    \mathrm{\sigma}_{n}= \sum_{j=1}^{J}{\alpha_{j} \cdot \operatorname{Var}\left (X_{n,j}\right)} = 
    \sum_{j=1}^{J}{\alpha_{j} \cdot\mathbb{E}\left[(X_{n,j}-\mathbb{E}\left[X_{n,j}\right])^{2}\right]}
\end{equation}
where $X_{n,j}$ is a random variable denoting the $n$-th neuron's response for a sample belonging to the $j$-th class. $J$ is the number of classes.
$\alpha_{j} = \frac{z_j}{\sum_i z_i}$ is the prior probability of the $j$-th class  where $z_j$ is the sample amount of $j$-th class. Notice that $\alpha_{j}$ is not a hyper-parameter, and it only presents the importance of different classes.
With the NSR term defined for each individual neuron, the overall regularization is derived by applying NSR term to all neurons in the network as follows:
\begin{equation}
    \mathcal{L}_{S}=\sum_{n=1}^{N}{\lambda_{n} \mathrm{\sigma}_{n}}
\end{equation}
where $\mathcal{L}_{S}$ represents the NSR term for the entire network, and $N$ is the number of neurons in the network. 
$\lambda_{n}$ is the hyper-parameter to  
control the regularization intensity. In this paper, for practical simplicity, $\lambda$ is set as the same value for all neurons.
Adding the overall regularization term to the main training target, i.e., the cross entropy loss, the final regularized loss function can be written as:
\begin{equation}
\label{equ:overall_loss}
\mathcal{L}=\mathcal{L}_{C}+\mathcal{L}_{S}=\mathcal{L}_{C}+\lambda\sum_{n=1}^{N}{ \mathrm{\sigma}_{n}},
\end{equation}
where $\mathcal{L}_{C}$ represents cross entropy loss. 

\subsection{Practical Implementation}
\label{implementation}
\subsubsection{Mini-batch Training}
In order to use mini-batch training, we adapt our NSR method to do forward and backward propagation 
based on mini-batches of samples. 
To be specific, we first transform Eq. \eqref{equ:neuron_stable_reg} as:
\begin{equation}
\begin{split}
    \sigma_n &= \sum_{j=1}^J \alpha_j
    \left(\mathbb{E}\left[X_{n, j}^2\right] - \mathbb{E}^2 \left[X_{n, j}\right]\right) \\
    &= \mathbb{E}\left[\sum_{j=1}^J \alpha_j X_{n, j}^2\right] - \sum_{j=1}^J \alpha_j \mathbb{E}^2 \left[ X_{n, j}\right].
\end{split}
\end{equation}
The first term has the same form as typical loss functions, i.e., an expectation of the function of data samples, which can be easily generalized to a mini-batch training setting via estimating the expectation with a batch of samples. Although the second term can be estimated in the same way, the square operation magnifies the estimation error, especially when the batch size is not large enough.

To alleviate the problem while not introducing much computing overhead, we propose a memory queue-based estimation method that allows us to leverage more history samples for estimation without additional sampling and forward/backward computation. For each class, we record the number of samples and the summation values of the neurons' response within each batch. To further reduce the storage overhead, we only maintain these values of the latest $M$ batches by two $M$-length queues. 

More specifically, an element $c_{m,j}$ in the first memory queue is the count number of $j$-th class instances in the $m$-th batch, represented as:
\begin{equation}
c_{m,j}=\sum_{y_i \in Y_m} \delta\left(y_{i}=j\right),
\end{equation}
where $Y_m$ is the set of labels in the $m$-th batch, $y_i$ is the label of $i$-th sample, and $\delta$ is a characteristic function, i.e., $\delta(condition)=1$ if $condition$ is satisfied, otherwise $\delta(condition)=0$. 
An element ${s}_{m,j}^{(n)}$ in the second memory queue is the summation value of $n$-th neuron's response for the samples belonging to the $j$-th class within the $m$-th batch, represented as:
\begin{equation}
{s}_{m,j}^{(n)}=\sum_{x^{(n)}_i\in \mathcal{X}_{m}^{(n)}} \delta\left(y_{i}=j\right) \cdot x^{(n)}_{i},
\end{equation}
where $\mathcal{X}_{m}^{(n)}$ is the set of the $n$-th neuron's response within the $m$-th batch and $x_{i}^{(n)}$ is the $n$-th neuron's response of the $i$-th sample.
When a new batch is fed, the estimation of expectation $\mathbb{E} \left[ X_{n, j}\right]$ can be updated by the following steps:
\begin{equation}
\begin{split}
    C_j := C_j - c_{0,j} + c_{*,j},& \quad S^{(n)}_j := S^{(n)}_j - s^{(n)}_{0,j} + s^{(n)}_{*,j} \\
    \hat{\mathbb{E}} \left[ X_{n, j}\right] &:= S_j^{(n)} / C_j, 
\end{split}
\end{equation}
where $\hat{\mathbb{E}} \left[ X_{n, j}\right]$ is the estimation of expectation $\mathbb{E} \left[ X_{n, j}\right]$, $C_j = \sum_m c_{m,j}$, $S_j^{(n)} = \sum_m s^{(n)}_{m,j}$, and $c_{*,j}$, $s^{(n)}_{*,j}$, as new elements appended to the queues, represent the count number and the summation for the new batch. Based on this dynamic update method, the additional memory overhead is negligible, and we give a space complexity analysis in Appendix \ref{app:complexity}.

\subsubsection{Layer Selection criterion for Applying NSR}
\label{sec:Steadiness redundancy}
\begin{figure}
\centering
    \subfigure[NSR only applied on the second layer]{
    \begin{minipage}[b]{0.4\textwidth}
    \centering
    \includegraphics[width=1\textwidth]{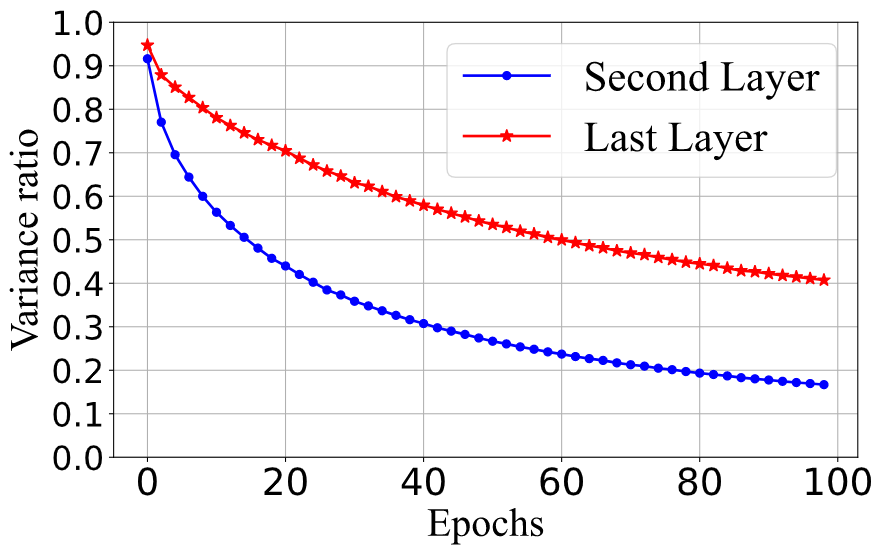}
    \end{minipage}
    }
    \hspace{0.6in}
    \subfigure[NSR only applied on the last layer]{
        \begin{minipage}[b]{0.4\textwidth}
        \centering
        \includegraphics[width=1\textwidth]{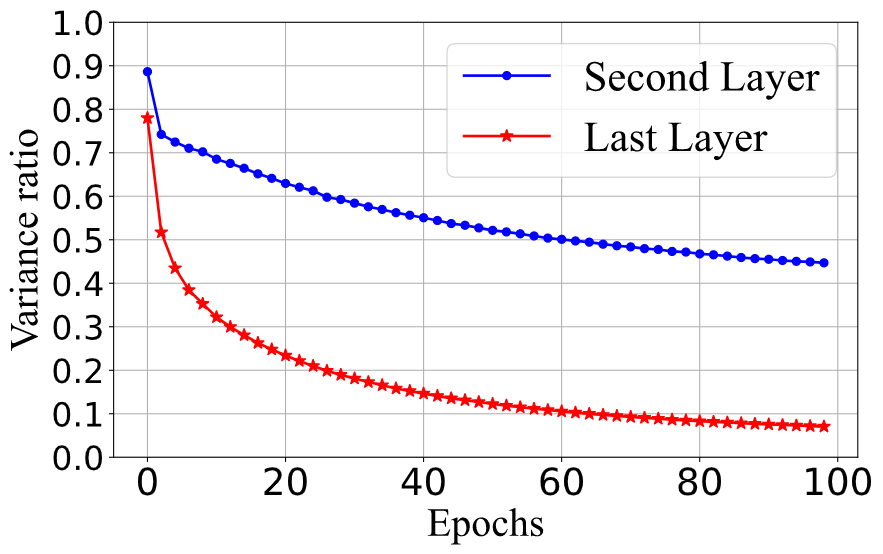}
        \end{minipage}
    }
    
    \caption{{The trend of variance ratio along with the training procedure on MNIST. The variance ratio is the neuron intra-class response variance of the three-layer MLP applied with NSR, (a) only on the second layer and (b) only on the last layer, divided by the corresponding variance of the vanilla three-layer MLP. }} 
  \label{fig:reduction}
\vspace{-0.4cm}
\end{figure}

We find that there is a correlation or redundancy among the steadiness constraints of different layers, meaning that applying NSR on different layers of the network has an overlapping effect on neuron steadiness control. In more detail, from Figure \ref{fig:reduction}, we can see that every layer's variance ratio decreases even only one specific layer is applied with NSR. Such steadiness correlation or redundancy among layers is not surprising because different layers are served as input/output of each other.

Due to the aforementioned redundancy, considering the trade-off between performance gain and computational overhead, we select only one layer to apply NSR. As our NSR is used to reduce neuron intra-class response variance, naturally, the intuitive criterion is to apply NSR to the layer with the largest aggregated neuron intra-class response variance. Detailed experiment results in RQ4 of the experiment section show such a criterion works well. In this paper, if not otherwise specified, we apply NSR to only one particular layer determined by this layer selection criterion.

\subsection{Theoretical Analysis}
\label{sec:theoretical_analysis}
In this section, we theoretically analyze the difference in model generalization ability with or without our proposed regularization method NSR based on Complexity Measure \cite{neyshabur2017exploring}.
The complexity Measure is one of the mainstream methods to measure the generalization ability of a deep learning model. A \textbf{lower complexity measure} means a \textbf{better generalization ability}. Formally, a Complexity Measure is a measure function $\mathcal{M}:\{\mathcal{H}, \mathcal{S}\}\to \mathbb{R}^+$ where $\mathcal{H}$ is a class of models and $\mathcal{S}$ is a training set. According to the definition, if a complexity measure of a given model tends to be 0, the model should have the best generalization ability. Note that, several different complexity measures have been proposed.

 We select Consistency of Representations \cite{natekar2020representation}
 as the concrete Complexity Measure used in our theoretical analysis. This measure is designed based on Davies-Bouldin Index \cite{davies1979cluster} and is the Winning Solution of the NeurIPS 2020 Competition on Predicting Generalization in Deep Learning. Mathematically, for a given dataset and a model, we define the following statistics:
\begin{align}
     S_i = \Big( \frac{1}{n_i} \sum^{n_i}_\tau | \mathcal{O}_i^{(\tau)} - \mu_{\mathcal{O}_i} |^p \Big)^{1/p} \; &{\rm for}\; i = 1 \cdots J \label{equ:intra-variance}\\
     M_{i, j} = || \mu_{\mathcal{O}_i} - \mu_{\mathcal{O}_j} ||_p \qquad &{\rm for}\; i,j = 1 \cdots J \label{equ:inter-variance},
\end{align}
where $i$ and $j$ are two different classes, and $n_i$ is the number of samples belonging to the class $i$. $\mathcal{O}_i^{(\tau)}$ is the output representation of the $\tau$-th sample belonging to class $i$ for the final layer, and $\mu_{\mathcal{O}_i}$ is the cluster centroid of the representations of class $i$. $S_i$ is a measure of scatter within representations of class $i$, and $M_{i, j}$ is a measure of separation between representations of classes $i$ and $j$. Then, the complexity measure based on the Davies-Bouldin Index is defined as:
\begin{equation}
\label{equ:complexity_measure}
    \mathcal{C} = \frac{1}{J} \sum_{i = 1}^{J} \max_{i \neq j} \frac{S_i + S_j}{M_{i, j}}.
\end{equation}

Based on the definition of the complexity measure $\mathcal{C}$, we have the following lemma:
\begin{lem}
     For a multi-class classification problem, when (1) a deep learning model utilizing the Cross Entropy loss with an NSR regularization term on any of its intermediate layers is optimized via gradient descent and (2) the capacity of the model is sufficiently large, the Consistency of Representations Complexity Measure on this model $\mathcal{C}$ will tend to be 0. Under the same condition, when the deep learning model is only optimized by the Cross Entropy loss without an NSR regularization term, there will be infinite local minima where the complexity measure $\mathcal{C}$ will be a positive number. 
\end{lem}
The proof is shown in Appendix \ref{app:proof}. Intuitively, this lemma shows that in an ideal condition (i.e., the model is sufficiently large), if the definition of the Consistency of Representations Complexity Measure is utilized, a deep learning model with NSR regularization term will be guaranteed to converge to local minima with the best generalization ability. In contrast, the model without NSR will not have the same guarantee.

\section{Experiment}
In this section, we conduct experiments over a variety of datasets to examine the performance of our \normethod on three extensively used neural network architectures. 
We design a series of experiments to answer the following research questions.  
\textbf{RQ1}: How does NSR perform on various datasets and different neural architectures? \textbf{RQ2}: Does NSR outperform other classical regularization methods? \textbf{RQ3}: What is the effect of combining NSR with other popular methods like Batch Normalization or Dropout? \textbf{RQ4}: What if we apply NSR to multiple layers instead of one particular layer?


\subsection{Experiment Setup}
\label{sec:experiment_settings}
\noindent{\textbf{Network Architectures and Datasets}.} 
Three architectures utilized in experiments are Multilayer Perceptron (MLP), Convolutional Neural Network (CNN) and Graph Neural Network (GNN). 
For vanilla models used as our baselines in different architectures, we adopt ResNet-18 \cite{he2016deep}, VGG-19 \cite{liu2018rethinking} and ResNet-50 for CNN, GCN \cite{kipf2016semi} and GraphSAGE \cite{hamilton2017inductive} for GNN. 
We run five MLP models which are denoted as MLP-L, L indicates the number of layers including the input layer. We elaborate on the details of these vanilla models in Appendix \ref{app:model_architecture}.
As for benchmark datasets used in our experiments,  MLP and CNN are applied to image recognition task on \textbf{MNIST} \cite{lecun1998mnist}, \textbf{CIFAR-10} \cite{krizhevsky2009learning} and \textbf{ImageNet} \cite{deng2009imagenet} datasets, respectively. GNN is applied to node classification on four real-world graph datasets: \textbf{WikiCS} \cite{mernyei2020wiki}, \textbf{PubMed} \cite{yang2016revisiting}, \textbf{Amazon-Photo} and \textbf{Amazon-Computers} \cite{shchur2018pitfalls}.  Notice that ImageNet is a large benchmark dataset with 1000 classes. Details are shown in Appendix \ref{app:dataset}.

\begin{table*}[!ht]
\centering
\caption{Error rate of applying our NSR on five MLP models for MNIST.}
\vspace{-0.2cm}
{%
\begin{tabular}{c|ccccc}
\toprule
Model     & MLP-3              & MLP-4              & MLP-6              & MLP-8              & MLP-10                        \\ \midrule
Vanilla (\%)  & 3.09   $\pm$ 0.10 & 2.29   $\pm$ 0.07 & 2.44   $\pm$ 0.09 & 2.87   $\pm$ 0.09 & 3.06   $\pm$ 0.06 \\
Vanilla+NSR (\%) & \textbf{2.80  $\pm$ 0.08}  & \textbf{1.64  $\pm$ 0.04}  & \textbf{1.76  $\pm$ 0.06}  & \textbf{1.98  $\pm$ 0.09} & \textbf{1.72 $\pm$ 0.14}\\
Gain      & 9.39\%             & 28.38\%            & 27.87\%            & 30.87\%            & 43.79\%                      \\ \bottomrule
\end{tabular}%
}
\label{tab:mlp}
\end{table*}

\begin{wraptable}{r}{0.55\textwidth}
\renewcommand\tabcolsep{2pt}
 \caption{Error rate of applying our NSR on ResNet-18 and VGG-19 for CIFAR-10, and top-5 error rate on ResNet-50 for ImageNet.}
 \vspace{-0.2cm}
{%
\begin{tabular}{c|ccc}
\toprule
Model     &  ResNet-18        & VGG-19  & ResNet-50 \\ \midrule
Vanilla  & 4.22 $\pm$ 0.07 & 9.19 $\pm$ 0.18 & 7.82 $\pm$ 0.07\\
Vanilla+NSR & \textbf{3.74 $\pm$ 0.08} & \textbf{8.09 $\pm$ 0.17} & \textbf{6.98 $\pm$ 0.08} \\
Gain      & 11.37\%  & 11.97\% & 10.74\%      \\ \bottomrule
\end{tabular}
\vspace{-0.2cm}
}

\label{tab:cnn}
\end{wraptable}

\noindent{\textbf{Experiment Settings}.}
We follow the typical implementation settings to conduct our experiments. 
For ResNet-18 and VGG-19 on CIFAR-10, we follow the detailed setting of \cite{zhang2019lookahead, gouk2018maxgain}. 
For ResNet-50 on ImageNet, we follow the official implementation provided by torchversion library \footnote{https://pytorch.org/hub/pytorch\_vision\_resnet/}.
For GraphSAGE and GCN, we follow the implementation setting of \cite{10.1145/3442381.3449896}.
For MNIST dataset, we divide 60000 training images into the training set with 50000 samples and the validation set with the remaining 10000 samples to select hyper-parameter. For each of the four graph datasets, it is randomly split into training, validation, and testing sets with a ratio of 6:2:2. 
Note that, SGD \cite{ruder2016overview} is used to optimize MLP, and Adam \cite{kingma2014adam} for other models except for ResNet-18 that is optimized by Momentum \cite{ruder2016overview} according to the implementation setting of \cite{zhang2019lookahead}.
We use the typical setting of batch size as 100 for all experiments.
To ensure the model convergence, training epochs are set as 100 for both MLP and GNN, 200 for ResNet and 500 for VGG.  

In all experiments except for RQ4, we apply NSR to only one particular layer with the same $\lambda$ for each neuron.
For this only one hyper-parameter $\lambda$, like most other regularization methods, we apply a random search strategy to find its proper value ranging from 1e-2 to 10.
The error rate is the evaluation metric and each result is averaged over 5 runs with different random seeds. The hardware environments are detailed in Appendix \ref{app:hard}.

\subsection{Experiment Results}
\textbf{RQ1: Performance of NSR}: 
Here, we discuss the performance of NSR over different models and present the results on Tab. \ref{tab:mlp} - \ref{tab:gnn} ``Gain'' is the percentage of relative reduction in the error rate. Tab. \ref{tab:mlp} demonstrates that NSR can improve the performance of MLP with different layers: the relative error rate is reduced by 9.39\% at least and 43.79\% at most. Besides, as the number of network layers increases, the gain of NSR shows a roughly upward trend. The four-layer MLP achieves the lowest classification error rate among all vanilla version baselines, and the accuracy becomes worse as the networks grow deeper. It indicates that deep MLP encounters a severe overfitting problem. The success of our regularization in addressing such a problem reveals the importance of stabilizing the response of each individual neuron to instances from the same class. 

For CNN models, Tab. \ref{tab:cnn} demonstrates that NSR can reduce the relative error rate for CIFAR-10 by 11.97\% on VGG-19 and 11.37\% on Resnet-18, and reduce the relative error rate (top-5) for ImageNet by 10.74\% on ResNet-50. It is worth mentioning that VGG-19, ResNet-18 and ResNet-50 have already adopted Batch Normalization, Dropout regularization and Weight Decay in vanilla models, the accuracy gain of our method reveals that NSR can have extra benefits to ulteriorly enhance generalization ability when combined with Batch Normalization, Dropout and Weight Decay. We will show more evidence about this in RQ3.

\begin{table*}[!ht]
\centering
\renewcommand\tabcolsep{2pt}
\caption{Error rate of applying our NSR on GCN and GraphSAGE over four graph datasets.}
\vspace{-0.2cm}
\begin{tabular}{l|c|cc|cc}
\toprule
Dataset & Layers & GraphSAGE (\%) & GraphSAGE+NSR (\%) & GCN (\%)   & GCN+NSR (\%) \\
\midrule
\multirow{3}{*}{PubMed} 
& 2 & 10.73 $\pm$ 0.06 & \textbf{9.89  $\pm$ 0.08} & 12.02 $\pm$ 0.00 & \textbf{11.92 $\pm$ 0.00} \\
& 3 & 10.20 $\pm$ 0.25 & \textbf{9.48  $\pm$ 0.12} & 12.76 $\pm$ 0.18 & \textbf{12.19 $\pm$ 0.11} \\
& 4 & 10.43 $\pm$ 0.17 & \textbf{9.79  $\pm$ 0.19} & 14.01 $\pm$ 0.07 & \textbf{12.96 $\pm$ 0.08} \\
\midrule
\multirow{3}{17mm}{Amazon-Photo} 
& 2 & 5.82  $\pm$ 0.00 & \textbf{4.54  $\pm$ 0.10} & 6.73  $\pm$ 0.00 & \textbf{6.27  $\pm$ 0.00} \\
& 3 & 5.20  $\pm$ 0.14 & \textbf{4.86  $\pm$ 0.13} & 8.00  $\pm$ 0.11 & \textbf{7.96  $\pm$ 0.10} \\
& 4 & 6.37  $\pm$ 0.30 & \textbf{5.62  $\pm$ 0.59} & 10.24 $\pm$ 0.14 & \textbf{9.03  $\pm$ 0.25} \\
\midrule
\multirow{3}{17mm}{Amazon-Computers} 
& 2 & 11.37 $\pm$ 0.55 & \textbf{10.47 $\pm$ 0.05} & 12.17 $\pm$ 0.07 & \textbf{10.86 $\pm$ 0.03} \\
& 3 & 11.88 $\pm$ 1.05 & \textbf{10.22 $\pm$ 0.54} & 14.90 $\pm$ 0.25 & \textbf{13.66 $\pm$ 0.12} \\
& 4 & 15.49 $\pm$ 0.90 & \textbf{12.86 $\pm$ 0.82} & 18.07 $\pm$ 0.74 & \textbf{16.02 $\pm$ 0.23} \\
\midrule
\multirow{3}{*}{WikiCS} 
& 2 & 16.81 $\pm$ 0.21 & \textbf{16.06 $\pm$ 0.33} & 18.41 $\pm$ 0.06 & \textbf{17.99 $\pm$ 0.05} \\
& 3 & 15.97 $\pm$ 0.18 & \textbf{15.27 $\pm$ 0.21} & 18.66 $\pm$ 0.23 & \textbf{18.10 $\pm$ 0.27} \\
& 4 & 16.63 $\pm$ 0.31 & \textbf{15.43 $\pm$ 0.24} & 19.21 $\pm$ 0.31 & \textbf{18.84 $\pm$ 0.26}\\
\bottomrule
\end{tabular}
\vspace{-0.2cm}
\label{tab:gnn}
\end{table*}

\begin{table*}

\begin{floatrow}
\centering
\renewcommand\tabcolsep{2pt}

\capbtabbox[6.9cm]{
 \begin{tabular}{c|ccc}
\toprule
 & MLP-4  &ResNet-18 &GraphSAGE      \\ \midrule
Vanilla & 2.29 $\pm$ 0.07& 7.96 $\pm$ 0.12& 11.37 $\pm$ 0.55 \\ 
L1       & 2.27 $\pm$ 0.05& 7.83 $\pm$ 0.23& 10.81 $\pm$ 0.13 \\ 
L2       & 2.27 $\pm$ 0.05& 7.67 $\pm$ 0.18& 10.68 $\pm$ 0.35 \\
Jacobian & 2.21 $\pm$ 0.04& 7.90 $\pm$ 0.07& 11.27 $\pm$ 0.45 \\ 
NSR       & \textbf{1.64 $\pm$ 0.04}& \textbf{7.20 $\pm$ 0.09}& \textbf{10.52 $\pm$ 0.22} \\ \bottomrule
\end{tabular}
}{
    \caption{Error rate comparison of different regularization methods on different models.}
    \label{tab:regularization}
}
\capbtabbox[6.9cm]{
 \begin{tabular}{c|ccc}
 \specialrule{0em}{1.7pt}{1.7pt}
\toprule
MLP-4 & Vanilla & + BN & + BN\&NSR     \\ \midrule
Error rate & 2.29 $\pm$ 0.07& 2.22 $\pm$ 0.04& \textbf{1.62 $\pm$ 0.08}         \\ \midrule
MLP-4 & Vanilla  &+ DO &+ DO\&NSR      \\ \midrule
Error rate & 2.29 $\pm$ 0.07& 2.19 $\pm$ 0.04& \textbf{1.64 $\pm$ 0.04}         \\ \bottomrule
\end{tabular}
}{
    \caption{Combination of our NSR with Batch Normalization (BN) and Dropout (DO) for training.}
    \label{tab:further gain}
}
\end{floatrow}

\end{table*}

\begin{table}[!h]
\caption{Effect of applying NSR on different layer(s) of MLP-4. The subscript number indicates which layer(s) NSR is applied on.}

\centering
\begin{tabular}{c|ccccc}
\toprule
            & MLP  & MLP$_2$  & MLP$_3$ &  MLP$_4$ & MLP$_{3,4}$    \\ \midrule

Error rate  & 2.29 $\pm$ 0.07& 2.22 $\pm$ 0.08& 1.90  $\pm$ 0.13 &  1.64 $\pm$ 0.04 & \textbf{1.63  $\pm$ 0.08}\\ \bottomrule
\end{tabular}
\label{tab:different_layer_MLP}
\vspace{-0.2cm}
\end{table}

\begin{table}[!h]
\caption{Effect of applying NSR on different layer(s) of GraphSAGE-2. The subscript number indicates which layer(s) NSR is applied on.}
\centering
\begin{tabular}{c|cccc}
\toprule
            & GraphSAGE  & GraphSAGE$_1$   & GraphSAGE$_2$  &  GraphSAGE$_{1,2}$    \\ \midrule

Error rate  & 11.37  $\pm$ 0.55& 10.47  $\pm$ 0.05 &  10.52   $\pm$ 0.22 &  \textbf{10.30   $\pm$ 0.17} \\ \bottomrule
\end{tabular}
\label{tab:different_layer_GNN}
\vspace{-0.4cm}
\end{table}

The number of layers of two GNN models varies from 2 to 4. It is a typical setting following empirical experiences as nodes will capture similar information from neighbors and result in over-smoothness when GNNs grow deeper. Tab. \ref{tab:gnn} shows that GraphSAGE and GCN applied with NSR outperform the vanilla model with different layer depths on all datasets. Specifically, GraphSAGE and GCN achieve an average improvement of 8.6\% and 5.8\%, respectively, and 17.0\% improvement at most.

\textbf{RQ2: Comparison with Other Regularization}: The vanilla ResNet18 here does not use any regularization term and learning rate decay methods for fairness. We first compare our NSR method with several classical regularization methods shown in Tab. \ref{tab:regularization}. Notice that, both our NSR and the other regularization methods for comparison have only one hyper-parameter to tune, and we utilize the same hyper-parameter searching strategy to select the best hyper-parameter values for 
all of them. As shown in Tab. \ref{tab:regularization}, our NSR performs best among these regularization methods on all three models with different architectures, i.e., MLP-4 (MLP), ResNet-18 (CNN), GraphSAGE (GNN). The results indicate that our NSR has remarkable improvement and is general for various architectures. 

\textbf{RQ3: Combination with Other Regularization}:
We further investigate the effect of combining our NSR with other popular methods like Batch Normalization and Dropout. We conduct the experiments on MLP-4 and the result are organized in Tab. \ref{tab:further gain}. From Tab. \ref{tab:further gain} we can find that both Batch Normalization and Dropout can reduce error rate compared with vanilla baseline, and adding our NSR on top of them can further promote the performance significantly. It indicates that NSR can provide complementary regularization benefits with Batch Normalization and Dropout.

\textbf{RQ4: Layer selection for applying NSR\label{sec: ablation}}: 
We have introduced how to select one particular layer to apply NSR in section \ref{sec:Steadiness redundancy}. To evaluate such layer selection criterion, we compare the performances trained by applying NSR to multiple layers or to only one selected layer, respectively. Taking MLP-4 on MNIST and GraphSAGE-2 on Amazon-Computers as two examples, we apply NSR to different layer(s), and their corresponding error rates are listed in Tab. \ref{tab:different_layer_MLP} $\sim$ Tab. \ref{tab:different_layer_GNN}, where model$_{l}$ means NSR is applied to the $l_{th}$ layer of the model. Notice that, except the first layer of MLP-4, the input layer, for the second, third, and last layer of MLP-4, their neuron intra-class response variance are 409, 510, and 1660, respectively. For GraphSAGE-2, its neuron intra-class response variance are 4.15 and 2.68, for the first and last layer, respectively. The variance of MLP-4 and GraphSAGE-2 are quite different because of the data characteristic difference between MNIST and Amazon-Computers. 

The results in Tab. \ref{tab:different_layer_MLP} - \ref{tab:different_layer_GNN} show that, no matter which layer(s) is applied with NSR, it could always improve the accuracy compared with the vanilla baseline for both MLP-4 and GraphSAGE-2.
In addition, according to our layer selection criterion, applying NSR to the layer with the biggest variance, i.e., the last layer for MLP-4 and the first layer for GraphSAGE-2, could achieve the most significant gain compared with applying NSR to other individual layers. Also, it could achieve similar accuracy compared with the best one obtained by applying NSR to multiple layers. This is the empirical evidence to demonstrate the rationality of our layer selection criterion.

\section{Related Work}
Regularization improves generalization ability by introducing the inductive bias based on prior knowledge. According to the type of prior knowledge, existing works can be roughly categorized into Domain Specific Regularization and Model Generic Regularization. 

\textbf{Domain Specific Regularization} shows success in various domains including Face Recognition \cite{wen2016discriminative, liu2017sphereface, cai2018island, liu2017adaptive},  Knowledge Graph Completion \cite{zhang2020duality,  lacroix2018canonical, minervini2017regularizing}, Graph Neural Network \cite{chen2020measuring,hou2019measuring,rong2019dropedge,yang2020domain,long2019hierarchical}.
In Face Recognition, \cite{cai2018island} points out the existence of external factors, such as different situations of environmental illumination, head poses, and facial expressions, which brings the challenge that faces images from the same person may have even larger differences than face images from different persons. To address specific challenges in face recognition, \cite{liu2017sphereface} imposes the carefully-designed regularization method which enforces the representations of face instances from the same person to be similar. 
In Graph Neural Network, the popular model usually suffers from the over-smoothing problem due to the six degrees of separation \cite{kleinfeld2002could}. \cite{chen2020measuring} statistically analyzes the high correlation between smoothness and the mean average distance (MAD) among node representation. Then a regularization called MADGap is proposed which punishes over smoothness by minimizing MAD. 

\textbf{Model Generic Regularization} can be categorized into network-wise Regularization, layer-wise Regularization, and neuron-wise Regularization, according to the granularity of the studied properties.
Network-wise Regularization encodes the desired property of the entire network, like the sparseness of the network and the smoothness of the mapping function between input and output. L2 Regularization \cite{plaut1986experiments, lang1990dimensionality} encourages small sum squared magnitude of model weights. 
\cite{hoffman2019robust} introduces an efficient framework to minimize the norm of the input-out Jacobian matrix for noise robustness.
\cite{foret2020sharpness} encourages model parameters to be uniformly low at some local regions of the loss function.
Layer regularization \cite{ulyanov2016instance, plaut1986experiments} becomes popular due to the success achieved by Batch Normalization \cite{ioffe2015batch}. 
Neuron-wise Regularization is the method with the most fine-grained granularity. Dropout \cite{hinton2012improving,srivastava2014dropout} is the typical example belonging to this category. It randomly removes neurons with a certain probability to avoid strong co-adaption between neurons. 
Activation Regularization \cite{merity2017revisiting} adds the L2 Regularization on masked activations of neurons and Temporal Activation Regularization restricts the difference between RNN outputs at adjacent timesteps. To minimize the variance of sample variance and obtain a few distinct modes, Variance Constancy Loss \cite{littwin2018regularizing} is applied on all neurons before the activation as regularization. 
Our proposed Neuron Steadiness Regularization belongs to the neuron-wise regularization that leverages information of individual neuron response distribution. 

\section{Conclusion and Future Work}
\label{sec:conclusion}
We explore the inductive bias from the new perspective of class-dependent response distribution of individual neurons. Based on experimental observations, we articulate the Neuron Steadiness Hypothesis and propose the Neuron Steadiness Regularization that penalizes large intra-class neuron response variance. Based on the Complexity Measure, we provide a theoretical analysis of its effectiveness for improving generalization. Additionally, we conduct extensive evaluations on diverse datasets with various network architectures to demonstrate its power. 
Especially, we demonstrate its effectiveness on the classification task with large models like ResNet-50, and large datasets with many classes like ImageNet with 1000 classes. We systematically consider the border impact, and No risk is found.


\bibliographystyle{plain}
\bibliography{ref}

\begin{thebibliography}{10}

\bibitem{brock2016neural}
Andrew Brock, Theodore Lim, James~M. Ritchie, and Nick Weston.
\newblock Neural photo editing with introspective adversarial networks.
\newblock In {\em Proceedings of International Conference on Learning
  Representations}, 2017.

\bibitem{cai2018island}
Jie Cai, Zibo Meng, Ahmed{-}Shehab Khan, Zhiyuan Li, James O'Reilly, and Yan
  Tong.
\newblock Island loss for learning discriminative features in facial expression
  recognition.
\newblock In {\em {IEEE} International Conference on Automatic Face {\&}
  Gesture Recognition}, 2018.

\bibitem{chen2020measuring}
Deli Chen, Yankai Lin, Wei Li, Peng Li, Jie Zhou, and Xu~Sun.
\newblock Measuring and relieving the over-smoothing problem for graph neural
  networks from the topological view.
\newblock In {\em Proceedings of the AAAI Conference on Artificial
  Intelligence}, pages 3438--3445, 2020.

\bibitem{cui2020towards}
Shuhao Cui, Shuhui Wang, Junbao Zhuo, Liang Li, Qingming Huang, and Qi~Tian.
\newblock Towards discriminability and diversity: Batch nuclear-norm
  maximization under label insufficient situations.
\newblock In {\em Proceedings of the IEEE/CVF Conference on Computer Vision and
  Pattern Recognition}, pages 3941--3950, 2020.

\bibitem{davies1979cluster}
David~L Davies and Donald~W Bouldin.
\newblock A cluster separation measure.
\newblock {\em IEEE transactions on pattern analysis and machine intelligence},
  (2):224--227, 1979.

\bibitem{deng2009imagenet}
Jia Deng, Wei Dong, Richard Socher, Li-Jia Li, Kai Li, and Li~Fei-Fei.
\newblock Imagenet: A large-scale hierarchical image database.
\newblock In {\em 2009 IEEE conference on computer vision and pattern
  recognition}, pages 248--255. Ieee, 2009.

\bibitem{foret2020sharpness}
Pierre Foret, Ariel Kleiner, Hossein Mobahi, and Behnam Neyshabur.
\newblock Sharpness-aware minimization for efficiently improving
  generalization.
\newblock {\em arXiv preprint arXiv:2010.01412}, 2020.

\bibitem{goodfellow2016deep}
Ian Goodfellow, Yoshua Bengio, and Aaron Courville.
\newblock {\em Deep learning}.
\newblock MIT press Cambridge, 2016.

\bibitem{gouk2018maxgain}
Henry Gouk, Bernhard Pfahringer, Eibe Frank, and Michael~J Cree.
\newblock Maxgain: Regularisation of neural networks by constraining activation
  magnitudes.
\newblock In {\em Joint European conference on machine learning and knowledge
  discovery in databases}, pages 541--556. Springer, 2018.

\bibitem{hamilton2017inductive}
William~L. Hamilton, Zhitao Ying, and Jure Leskovec.
\newblock Inductive representation learning on large graphs.
\newblock In {\em Proceedings of NuerIPS}, 2017.

\bibitem{he2016deep}
Kaiming He, Xiangyu Zhang, Shaoqing Ren, and Jian Sun.
\newblock Deep residual learning for image recognition.
\newblock In {\em Proceedings of the IEEE Conference on Computer Vision and
  Pattern Recognition}, pages 770--778, 2016.

\bibitem{hinton2012improving}
Geoffrey~E Hinton, Nitish Srivastava, Alex Krizhevsky, Ilya Sutskever, and
  Ruslan~R Salakhutdinov.
\newblock Improving neural networks by preventing co-adaptation of feature
  detectors.
\newblock {\em arXiv preprint arXiv:1207.0580}, 2012.

\bibitem{hoffman2019robust}
Judy Hoffman, Daniel~A Roberts, and Sho Yaida.
\newblock Robust learning with jacobian regularization.
\newblock {\em arXiv preprint arXiv:1908.02729}, 2019.

\bibitem{hou2019measuring}
Yifan Hou, Jian Zhang, James Cheng, Kaili Ma, Richard~TB Ma, Hongzhi Chen, and
  Ming-Chang Yang.
\newblock Measuring and improving the use of graph information in graph neural
  networks.
\newblock In {\em Proceedings of International Conference on Learning
  Representations}, 2019.

\bibitem{ioffe2015batch}
Sergey Ioffe and Christian Szegedy.
\newblock Batch normalization: Accelerating deep network training by reducing
  internal covariate shift.
\newblock In {\em Proceedings of International Conference on Machine Learning},
  pages 448--456, 2015.

\bibitem{kingma2014adam}
Diederik~P Kingma and Jimmy Ba.
\newblock Adam: A method for stochastic optimization.
\newblock In {\em Proceedings of International Conference on Learning
  Representations}, 2015.

\bibitem{kipf2016semi}
Thomas~N. Kipf and Max Welling.
\newblock Semi-supervised classification with graph convolutional networks.
\newblock {\em In Proceedings of International Conference on Learning
  Representations}, 2017.

\bibitem{kleinfeld2002could}
Judith Kleinfeld.
\newblock Could it be a big world after all? the six degrees of separation
  myth.
\newblock {\em Society, April}, 12:5--2, 2002.

\bibitem{krizhevsky2009learning}
Alex Krizhevsky, Geoffrey Hinton, et~al.
\newblock Learning multiple layers of features from tiny images.
\newblock 2009.

\bibitem{lacroix2018canonical}
Timoth{\'e}e Lacroix, Nicolas Usunier, and Guillaume Obozinski.
\newblock Canonical tensor decomposition for knowledge base completion.
\newblock In {\em International Conference on Machine Learning}, pages
  2863--2872, 2018.

\bibitem{lang1990dimensionality}
Kevin~J Lang and Geoffrey~E Hinton.
\newblock Dimensionality reduction and prior knowledge in e-set recognition.
\newblock In {\em Advances in Neural Information Processing Systems}, pages
  178--185, 1990.

\bibitem{lecun1998mnist}
Yann LeCun.
\newblock The mnist database of handwritten digits.
\newblock {\em http://yann. lecun. com/exdb/mnist/}, 1998.

\bibitem{littwin2018regularizing}
Etai Littwin and Lior Wolf.
\newblock Regularizing by the variance of the activations' sample-variances.
\newblock {\em Advances in Neural Information Processing Systems}, 31, 2018.

\bibitem{liu2017sphereface}
Weiyang Liu, Yandong Wen, Zhiding Yu, Ming Li, Bhiksha Raj, and Le~Song.
\newblock Sphereface: Deep hypersphere embedding for face recognition.
\newblock In {\em Proceedings of CVPR}, 2017.

\bibitem{liu2017adaptive}
Xiaofeng Liu, BVK Vijaya~Kumar, Jane You, and Ping Jia.
\newblock Adaptive deep metric learning for identity-aware facial expression
  recognition.
\newblock In {\em Proceedings of the IEEE Conference on Computer Vision and
  Pattern Recognition Workshops}, pages 20--29, 2017.

\bibitem{liu2018rethinking}
Zhuang Liu, Mingjie Sun, Tinghui Zhou, Gao Huang, and Trevor Darrell.
\newblock Rethinking the value of network pruning.
\newblock In {\em Proceedings of International Conference on Learning
  Representations}, 2018.

\bibitem{long2019hierarchical}
Qingqing Long, Yiming Wang, Lun Du, Guojie Song, Yilun Jin, and Wei Lin.
\newblock Hierarchical community structure preserving network embedding: A
  subspace approach.
\newblock In {\em Proceedings of the 28th ACM International Conference on
  Information and Knowledge Management}, pages 409--418, 2019.

\bibitem{10.1145/3442381.3449896}
Xiaojun Ma, Junshan Wang, Hanyue Chen, and Guojie Song.
\newblock Improving graph neural networks with structural adaptive receptive
  fields.
\newblock In {\em Proceedings of the Web Conference 2021}, page 2438–2447,
  2021.

\bibitem{merity2017revisiting}
Stephen Merity, Bryan McCann, and Richard Socher.
\newblock Revisiting activation regularization for language rnns.
\newblock {\em arXiv preprint arXiv:1708.01009}, 2017.

\bibitem{mernyei2020wiki}
P{\'e}ter Mernyei and C{\u{a}}t{\u{a}}lina Cangea.
\newblock Wiki-cs: A wikipedia-based benchmark for graph neural networks.
\newblock {\em arXiv preprint arXiv:2007.02901}, 2020.

\bibitem{minervini2017regularizing}
Pasquale Minervini, Luca Costabello, Emir Mu{\~n}oz, V{\'\i}t
  Nov{\'a}{\v{c}}ek, and Pierre-Yves Vandenbussche.
\newblock Regularizing knowledge graph embeddings via equivalence and inversion
  axioms.
\newblock In {\em Proceedings of Joint European Conference on Machine Learning
  and Knowledge Discovery in Databases}, pages 668--683, 2017.

\bibitem{miotto2018deep}
Riccardo Miotto, Fei Wang, Shuang Wang, Xiaoqian Jiang, and Joel~T Dudley.
\newblock Deep learning for healthcare: review, opportunities and challenges.
\newblock {\em Briefings in bioinformatics}, 19(6):1236--1246, 2018.

\bibitem{natekar2020representation}
Parth Natekar and Manik Sharma.
\newblock Representation based complexity measures for predicting
  generalization in deep learning.
\newblock {\em arXiv preprint arXiv:2012.02775}, 2020.

\bibitem{naumov2019deep}
Maxim Naumov, Dheevatsa Mudigere, Hao-Jun~Michael Shi, Jianyu Huang, Narayanan
  Sundaraman, Jongsoo Park, Xiaodong Wang, Udit Gupta, Carole-Jean Wu,
  Alisson~G Azzolini, et~al.
\newblock Deep learning recommendation model for personalization and
  recommendation systems.
\newblock {\em arXiv preprint arXiv:1906.00091}, 2019.

\bibitem{neyshabur2017exploring}
Behnam Neyshabur, Srinadh Bhojanapalli, David McAllester, and Nathan Srebro.
\newblock Exploring generalization in deep learning.
\newblock In {\em Proceedings of the 31st International Conference on Neural
  Information Processing Systems}, pages 5949--5958, 2017.

\bibitem{plaut1986experiments}
David~C Plaut et~al.
\newblock Experiments on learning by back propagation.
\newblock 1986.

\bibitem{rao2018deep}
Qing Rao and Jelena Frtunikj.
\newblock Deep learning for self-driving cars: Chances and challenges.
\newblock In {\em Proceedings of IEEE/ACM International Workshop on Software
  Engineering for AI in Autonomous Systems}, pages 35--38, 2018.

\bibitem{rawassizadeh2019manifestation}
Reza Rawassizadeh, Taylan Sen, Sunny~Jung Kim, Christian Meurisch, Hamidreza
  Keshavarz, Max M{\"u}hlh{\"a}user, and Michael Pazzani.
\newblock Manifestation of virtual assistants and robots into daily life:
  Vision and challenges.
\newblock {\em CCF Transactions on Pervasive Computing and Interaction},
  1(3):163--174, 2019.

\bibitem{rong2019dropedge}
Yu~Rong, Wenbing Huang, Tingyang Xu, and Junzhou Huang.
\newblock Dropedge: Towards deep graph convolutional networks on node
  classification.
\newblock In {\em Proceedings of International Conference on Learning
  Representations}, 2020.

\bibitem{ruder2016overview}
Sebastian Ruder.
\newblock An overview of gradient descent optimization algorithms.
\newblock {\em arXiv preprint arXiv:1609.04747}, 2016.

\bibitem{shchur2018pitfalls}
Oleksandr Shchur, Maximilian Mumme, Aleksandar Bojchevski, and Stephan
  G{\"u}nnemann.
\newblock Pitfalls of graph neural network evaluation.
\newblock {\em arXiv preprint arXiv:1811.05868}, 2018.

\bibitem{simonyan2014very}
Karen Simonyan and Andrew Zisserman.
\newblock Very deep convolutional networks for large-scale image recognition.
\newblock {\em arXiv preprint arXiv:1409.1556}, 2014.

\bibitem{sokolic2017robust}
Jure Sokoli{\'c}, Raja Giryes, Guillermo Sapiro, and Miguel~RD Rodrigues.
\newblock Robust large margin deep neural networks.
\newblock {\em IEEE Transactions on Signal Processing}, 65(16):4265--4280,
  2017.

\bibitem{srivastava2014dropout}
Nitish Srivastava, Geoffrey Hinton, Alex Krizhevsky, Ilya Sutskever, and Ruslan
  Salakhutdinov.
\newblock Dropout: a simple way to prevent neural networks from overfitting.
\newblock {\em The journal of machine learning research}, 15(1):1929--1958,
  2014.

\bibitem{ulyanov2016instance}
Dmitry Ulyanov, Andrea Vedaldi, and Victor Lempitsky.
\newblock Instance normalization: The missing ingredient for fast stylization.
\newblock {\em arXiv preprint arXiv:1607.08022}, 2016.

\bibitem{wen2016discriminative}
Yandong Wen, Kaipeng Zhang, Zhifeng Li, and Yu~Qiao.
\newblock A discriminative feature learning approach for deep face recognition.
\newblock In {\em Proceedings of European Conference on Computer Vision}, pages
  499--515, 2016.

\bibitem{yang2020domain}
Shuwen Yang, Guojie Song, Yilun Jin, and Lun Du.
\newblock Domain adaptive classification on heterogeneous information networks.
\newblock In {\em IJCAI}, pages 1410--1416, 2020.

\bibitem{yang2016revisiting}
Zhilin Yang, William Cohen, and Ruslan Salakhudinov.
\newblock Revisiting semi-supervised learning with graph embeddings.
\newblock In {\em Proceedings of International Conference on Machine Learning},
  pages 40--48, 2016.

\bibitem{zhang2019lookahead}
Michael~R Zhang, James Lucas, Geoffrey Hinton, and Jimmy Ba.
\newblock Lookahead optimizer: k steps forward, 1 step back.
\newblock {\em arXiv preprint arXiv:1907.08610}, 2019.

\bibitem{zhang2020duality}
Zhanqiu Zhang, Jianyu Cai, and Jie Wang.
\newblock Duality-induced regularizer for tensor factorization based knowledge
  graph completion.
\newblock {\em Proceedings of Advances in Neural Information Processing
  Systems}, 2020.

\end{thebibliography}
\section*{Checklist}


\begin{enumerate}

\item For all authors...
\begin{enumerate}
  \item Do the main claims made in the abstract and introduction accurately reflect the paper's contributions and scope?
    \answerYes{}
  \item Did you describe the limitations of your work?
    \answerYes{}
  \item Did you discuss any potential negative societal impacts of your work?
    \answerYes{}
  \item Have you read the ethics review guidelines and ensured that your paper conforms to them?
    \answerYes{}
\end{enumerate}

\item If you are including theoretical results...
\begin{enumerate}
  \item Did you state the full set of assumptions of all theoretical results?
    \answerYes{}
        \item Did you include complete proofs of all theoretical results?
    \answerYes{}
\end{enumerate}

\item If you ran experiments...
\begin{enumerate}
  \item Did you include the code, data, and instructions needed to reproduce the main experimental results (either in the supplemental material or as a URL)?
    \answerYes{We share the corresponding link in Appendix \ref{app:code}}
  \item Did you specify all the training details (e.g., data splits, hyperparameters, how they were chosen)?
    \answerYes{See in Appendix \ref{app:dataset}}
        \item Did you report error bars (e.g., with respect to the random seed after running experiments multiple times)?
    \answerYes{Yes, each results is calculated from five random seeds: [1, 3, 5, 7, 9]}
        \item Did you include the total amount of compute and the type of resources used (e.g., type of GPUs, internal cluster, or cloud provider)?
    \answerYes{See in Appendix \ref{app:hard}}
\end{enumerate}

\item If you are using existing assets (e.g., code, data, models) or curating/releasing new assets...
\begin{enumerate}
  \item If your work uses existing assets, did you cite the creators?
    \answerYes{}
  \item Did you mention the license of the assets?
    \answerYes{See in Appendix \ref{app:dataset}}
  \item Did you include any new assets either in the supplemental material or as a URL?
    \answerNA{}
  \item Did you discuss whether and how consent was obtained from people whose data you're using/curating?
    \answerNA{}
  \item Did you discuss whether the data you are using/curating contains personally identifiable information or offensive content?
    \answerNA{}
\end{enumerate}

\item If you used crowdsourcing or conducted research with human subjects...
\begin{enumerate}
  \item Did you include the full text of instructions given to participants and screenshots, if applicable?
    \answerNA{}
  \item Did you describe any potential participant risks, with links to Institutional Review Board (IRB) approvals, if applicable?
    \answerNA{}
  \item Did you include the estimated hourly wage paid to participants and the total amount spent on participant compensation?
    \answerNA{}
\end{enumerate}

\end{enumerate}

\newpage
\appendix

\newtheorem{innercustomgeneric}{\customgenericname}
\providecommand{\customgenericname}{}
\newcommand{\newcustomtheorem}[2]{%
  \newenvironment{#1}[1]
  {%
   \renewcommand\customgenericname{#2}%
   \renewcommand\theinnercustomgeneric{##1}%
   \innercustomgeneric
  }
  {\endinnercustomgeneric}
}
\newcustomtheorem{customlemma}{Lemma}

\section{Proof \label{app:proof}}
\subsection{Lemma 3.1}
\begin{customlemma}{3.1}
	  For a multi-class classification problem, when (1) a deep learning model utilizing the Cross Entropy loss with an NSR regularization term on any of its intermediate layer is optimized via gradient descent and (2) the capacity of the model is sufficiently large, the Consistency of Representations Complexity Measure on this model $\mathcal{C}$ will tend to be 0. Under the same condition, when the deep learning model is only optimized by the Cross Entropy loss without an NSR regularization term, there will be infinite local minima where the complexity measure $\mathcal{C}$ will be a positive number. 
\end{customlemma}
\begin{proof}
Let us consider the original Cross Entropy loss first:
\begin{equation}
    \mathcal{L}_{C} = -\sum_{\tau = 1}^{\mathcal{N}} \log q(y = y_\tau |x_\tau; \Theta),
\end{equation}
where $\mathcal{N}$ is the number of training samples, $y_\tau$ is the label of the $\tau_{th}$ sample, $x_\tau$ is the feature of the $\tau_{th}$ sample, $\Theta$ is the parameters of the model, and $q(y|x_\tau; \Theta)$ is the predicted probability modeled by the deep learning model. Considering a single sample, we have the Cross Entropy loss for the $\tau_{th}$ sample:
\begin{equation}
\begin{split}
    \mathcal{L}_{C}^{(\tau)} &= -\log q(y = y_\tau |x_\tau; \Theta) = - \log q_\tau(y),
\end{split}
\end{equation}
where $q_\tau(y) \triangleq q(y = y_\tau |x_\tau; \Theta)$ for brevity.
Because the deep learning models for the classification task always have a normalization operation (e.g., Softmax) to make the final unconstrained representation be a probability form, $q_\tau(y)$ has the following form:
\begin{equation}
    q_\tau(y) =  \frac{a_y^{(\tau)}}{\sum_{j=1}^J a_j^{(\tau)}},
\end{equation}
where $J$ is the number of classes, $\mathbf{a^{(\tau)}} = [a^{(\tau)}_1, ..., a^{(\tau)}_J]$ is the final representation for the $\tau$-th sample. 
The loss can be rewritten as:
\begin{equation}
    \mathcal{L}_{C}^{(\tau)} = - \log \frac{a_y^{(\tau)}}{A_\tau}
\end{equation}
where $A_{\tau} = \sum_{j=1}^J a_j^{(\tau)}$ is the normalization factor. Since we will use gradient descent to optimize the objective function, we calculate the partial derivatives of the objective for the representations $h^{(\tau)}_j$:
\begin{equation}
\begin{split}
    \frac{\partial \mathcal{L}_{C}^{(\tau)}}{\partial h_y^{(\tau)}} &= \frac{a_y^{(\tau)}}{A_\tau} - 1\\
    \frac{\partial \mathcal{L}_{C}^{(\tau)}}{\partial h_j^{(\tau)}} &= \frac{a_j^{(\tau)}}{A_\tau}, \; {\rm for} \; j: j \neq y.
\end{split}
\end{equation}

Let the gradient be zero, then we obtain the sufficient conditions of the local minima:
\begin{equation}
    \begin{split}
        a_j^{(\tau)} &= 0 \; {\rm for} \; j: j \neq y,\\
        A_\tau &= a_y,\\
        a_y^{(\tau)} &> 0,\\
        \sum_j a_j^{(\tau)} &= A_\tau
    \end{split}
\end{equation}
In that case, under the assumption that the model is sufficiently large, the converged local minima for the $\tau$-th sample are the vectors where only the $y_\tau$-th entry is an arbitrary positive real number and the other entries are 0. 

According to the definition of the Consistency of Representations Complexity Measure Eq. \eqref{equ:complexity_measure}, it is easy to see that we have an infinite number of local minima where the measures $\mathcal{S}_i$ and $\mathcal{M}_{i, j}$ for arbitrary classes $i$ and $j$ are finite positive numbers. Thus, the Consistency of Representations Complexity Measure $\mathcal{C}$ will be a positive number in those local minima.

Let us take our proposed NSR regularization term $\mathcal{L}_{S}$ (refer to Eq. \eqref{equ:overall_loss}) into consideration. When adding NSR to $k$-th layer, the loss of NSR can be rewritten as:
\begin{equation}
    \mathcal{L}_{S} = \sum_i^J \sum^{n_i}_\tau || \mathcal{O}_{i,k}^{(\tau)} - \mu_{\mathcal{O}_{i,k}} ||^2,
\end{equation}
where $n_i$ is the number of samples belonging to the class $i$, $\mathcal{O}_{i,k}^{(\tau)}$ is the $k$-th layer output representation of the $\tau$-th sample belonging to class $i$, $\mu_{\mathcal{O}_{i,k}}$ is the cluster centroid of the $k$-th layer representations of class $i$. If we set $\mathcal{O}_{i,k}^{(\tau)}$ as optimization variables, it is easy to see that the global minima is $\mathcal{O}_{i,k}^{(\tau)} = \mu_{\mathcal{O}_{i,k}}$. It means that the intermediate representations of the same class in layer $k$ will be the same vector. Combined with the conditions of the local minima of Cross Entropy loss, we have the optimal final representations of the overall loss, i.e., 
\begin{gather}
\mathbf{a}^{\tau}_i = 
\begin{bmatrix}
\smash[b]{\underbrace{0 \dots 0}_{\text{\tiny $i-1$ times }}}
C_i
\smash[b]{\underbrace{0 \dots 0}_{\text{\tiny $J-i$ times }}}
\end{bmatrix}
\;{\rm for}\; i = 1 \cdots J,
\end{gather}
~\\
where $\mathbf{a}^{\tau}_i$ is the final representation vector of the $\tau$-th sample belonging to the class $i$, and $C_i$ is a positive real number that satisfies $C_i \neq C_j$ if $i \neq j$. It means that the final representations for the samples belonging to the same class are the same, while the representations are different for the samples belonging to different classes.
It is obvious that in that case, $\mathcal{S}_i$ is zero and $\mathcal{M}_{i, j}$ for arbitrary classes $i$ and $j$ is a finite positive number. Thus, the Consistency of Representations Complexity Measure $\mathcal{C}$ will be 0.
The Lemma is proven.
\end{proof}

\section{Hardware and Software Environment}
\label{app:hard}
The experiments are performed on two Linux servers (CPU: Intel(R) Xeon(R) CPU E5-2690 v4 @ 2.60GHz, Operation system: Ubuntu 16.04.6 LTS). For GPU resources, two NVIDIA Tesla V100 cards are used for MLP and GNN and ResNet-50 on ImageNet
experiments while two Titan V cards are used for other CNN models. 
The python libraries we use to implement our experiments are PyTorch 1.7.1 and PyG 1.6.3. 
 
\section{Details of Reproduction}
\label{app:code}
\subsection{Open Source Code}
We publish our code in Github (i.e., \url{https://github.com/lundu28/NSR}).

\subsection{Details of Baseline Methods}
\label{app:model_architecture}
In this subsection, we detailed the network architectures and the baseline methods used in our experiments.

\noindent\textbf{MLP} Multilayer Perceptron is a fundamental deep neural network architecture that is composed of an input layer, several hidden layers and an output layer. Five MLP models with different numbers of hidden layers are used to test the performance of \normethod on MNIST for handwritten digit classification. ReLU is used as an activation function in the experiments. We denote MLP-L as an L-layer MLP network. For example, MLP-4 is a 4-layer MLP with 2 hidden layers.  

\noindent\textbf{CNN} Convolutional neural networks are widely used in the computer vision (CV) field. Two famous ones among them, ResNet-18 \cite{he2016deep}, VGG-19 \cite{liu2018rethinking}, are chosen as the baseline methods in our paper for the image recognition task on CIFAR-10 data set. ResNet-50 \cite{he2016deep} is chosen for the ImageNet dataset \cite{deng2009imagenet}. \textbf{ResNet-18} \cite{he2016deep} is a 20-layer convolutional neural network with residual connections to avoid the problem of vanishing gradient. \textbf{ResNet-50} is similar to ResNet-18 except that it is a more expressive 50-layer convolutional neural network.
\textbf{VGG-19} \cite{simonyan2014very} is another classic deep convolutional neural network architecture. We utilize a variant of VGG-19 adapted for CIFAR-10 \cite{liu2018rethinking} classification.
    
\noindent\textbf{GNN} Graph neural networks achieve fantastic results on graph structure data. GCN \cite{kipf2016semi} and GraphSAGE are two classical methods in classifying nodes on real network data. \textbf{GCN} \cite{kipf2016semi} is inspired by CNN and introduces spectral graph convolutions as the layer-wise propagation on graphs. \textbf{GraphSAGE} \cite{hamilton2017inductive} is the first inductive GNN framework that samples neighbor nodes on graphs. 

The detailed settings of different architectures of MLP and GNN are shown in Tab. \ref{tab:network structure}. Notice that we only show the architecture of hidden layers. The dimensions of the input and output layer rely on the input feature size and the number of categories of different datasets respectively.

\begin{table*}[!ht]
\centering
\caption{The detail architectures of MLP and GNN.}
\begin{tabular}{l|l}
\toprule
Model & Hidden layer dimension \\ \midrule
MLP-3 & [100]\\
MLP-4 & [256, 100] \\
MLP-6 & [256, 128, 64, 32] \\
MLP-8 & [256, 128, 64, 32, 32, 16] \\
MLP-10 & [256, 128, 64, 64, 32, 32, 16, 16] \\
GNN-2 & [100] \\
GNN-3 & [256, 100] \\
GNN-4 & [256, 128, 64] \\
\bottomrule
\end{tabular}
\label{tab:network structure}
\end{table*}

\begin{table*}[!ht]
    \centering
    \caption{The final hyperparameter settings $\lambda$  for MLP and CNN \label{tab:final_hyper}}
    \begin{tabular}{c|cccccccc}
    \toprule
        Model & MLP-3 & MLP-4& MLP-6& MLP-8 & MLP-10& ResNet-18 & VGG-19 & ResNet-50 \\ \midrule
        $\lambda$ & 0.025 & 7.487 & 0.054 & 0.254 & 0.416 & 0.05 & 0.05 & 0.40 \\ \bottomrule
    \end{tabular}
\end{table*}

\begin{table*}[!ht]
\centering
\caption{The final hyperparamter settings $\lambda$  for GNN.}
\begin{tabular}{l|c|c|c}
\toprule
Dataset & Layers & GraphSAGE & GCN \\
\midrule
\multirow{3}{*}{PubMed} 
& 2 &  0.3909 & 0.0367\\  
& 3 & 0.1436 & 0.1994\\
& 4 & 0.5485 & 0.8831\\
\midrule
\multirow{3}{17mm}{Amazon-Photo} 
& 2 & 0.0012& 0.0069\\
& 3 & 0.3485 & 0.0855\\
& 4 & 0.0001 & 0.0024\\
\midrule
\multirow{3}{17mm}{Amazon-Computers} 
& 2 & 0.0014 & 0.0025\\
& 3 & 0.0037& 0.0008\\
& 4 & 0.0021& 0.0028\\
\midrule
\multirow{3}{*}{WikiCS} 
& 2 & 0.0274& 0.0572\\
& 3 & 0.0142& 0.0669\\
& 4 & 0.2226& 0.0015\\
\bottomrule

\end{tabular}

\label{tab:final_hyper_gnn}
\end{table*}

\subsection{Hyper-parameter Settings}
We have only one unique hyperparameter, i.e., $\lambda$ in our methods, and the detailed settings are given in the Tab. \ref{tab:final_hyper} and Tab. \ref{tab:final_hyper_gnn}. The search methods and search range are given in the main text. 
All the other hyper-parameters are set the same as baselines.

\subsection{Details of Datasets}
\label{app:dataset}
In this section, we describe detailed information about the public available 
benchmark data sets used in our experiments.

\textbf{MNIST} \cite{lecun1998mnist} is a handwritten digits dataset containing 60000 training images and 10000 testing images. It is used to examine the performances of MLP with different hidden layers. \textbf{CIFAR-10} \cite{krizhevsky2009learning} is used for ResNet-18 and VGG-19, which consists of 50,000 32$\times$32 training color images and 10,000 testing images categorized as 10 classes. 
\textbf{ImageNet} is a benchmark dataset used for ResNet-50, which contains 14,197,122 annotated images with 1000 classes. 
For GNN, we selected four real-world graph datasets: \textbf{PubMed} \cite{yang2016revisiting} is a paper citation network where nodes represent documents and edges represent citation links. Both \textbf{Amazon-Photo} and \textbf{Amazon-Computers} \cite{shchur2018pitfalls} are subgraphs of Amazon co-purchase graph where nodes correspond to goods and edges connecting two nodes denote they are frequently bought together. \textbf{WikiCS} \cite{mernyei2020wiki} is derived from Wikipedia whose nodes correspond to computer science articles and edges are hyperlinks. The license and source of these datasets are shown in Tab. \ref{tab:license} and Tab. \ref{tab:link}, respectively. All the datasets are public and widely used in many research works. The collectors of the datasets are responsible for ensuring no privacy and consent issues.
\begin{table*}[!ht]
\centering
\caption{The license of datasets used in this paper.}
\begin{tabular}{c|c}
\toprule
Dataset & license \\ \midrule
MNIST & Creative Commons Attribution-Share Alike 3.0 license\\
CIFAR-10 & MIT license \\
ImageNet & MIT license \\
PubMed & NLM license \\
WikiCS & MIT license \\
Amazon-Photo & MIT license \\
Amazon-Computers & MIT license \\
\bottomrule
\end{tabular}
\label{tab:license}
\end{table*}

\begin{table*}[!ht]
\caption{The source of datasets used in this paper.}
\centering
\begin{tabular}{c|c}
\toprule
Dataset & link \\ \midrule
MNIST & http://yann.lecun.com/exdb/mnist/\\
CIFAR-10 & https://www.cs.toronto.edu/~kriz/cifar.html \\
ImageNet & https://image-net.org/index\\
PubMed & https://github.com/shchur/gnn-benchmark/raw/master/data/planetoid/\\
WikiCS & https://github.com/pmernyei/wiki-cs-dataset \\
Amazon-Photo & https://github.com/shchur/gnn-benchmark/raw/master/data/npz/ \\
Amazon-Computers & https://github.com/shchur/gnn-benchmark/raw/master/data/npz/ \\
\bottomrule
\end{tabular}
\label{tab:link}
\end{table*}
\begin{figure*}[!h]
    \centering

    \subfigure[Cross entropy loss]{
        \begin{minipage}[b]{0.45\textwidth}
        \centering
        \includegraphics[width=1\textwidth]{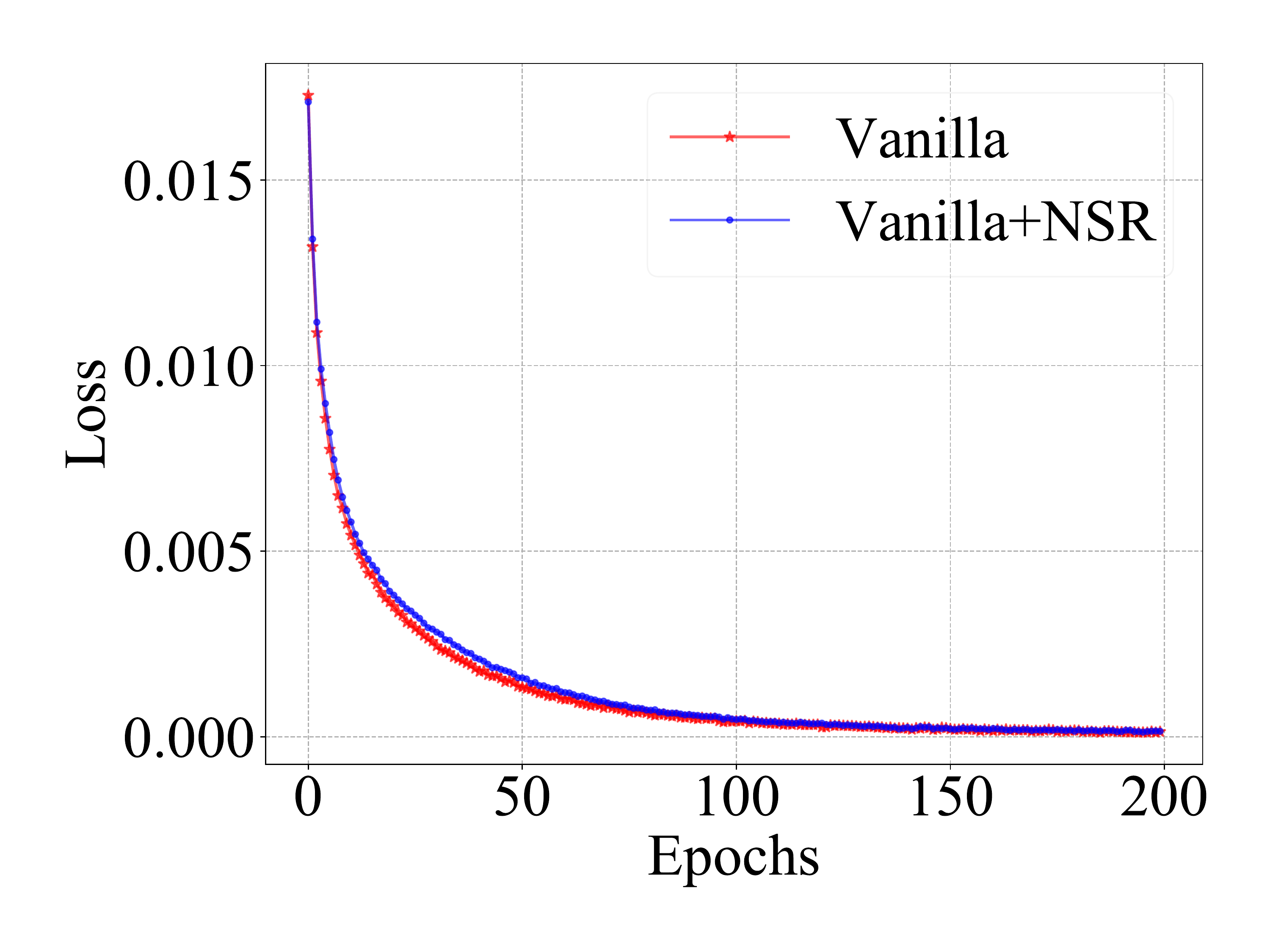}
        \end{minipage}
    }
    \subfigure[Intra-class variance]{
        \begin{minipage}[b]{0.45\textwidth}
        \centering
        \includegraphics[width=1\textwidth]{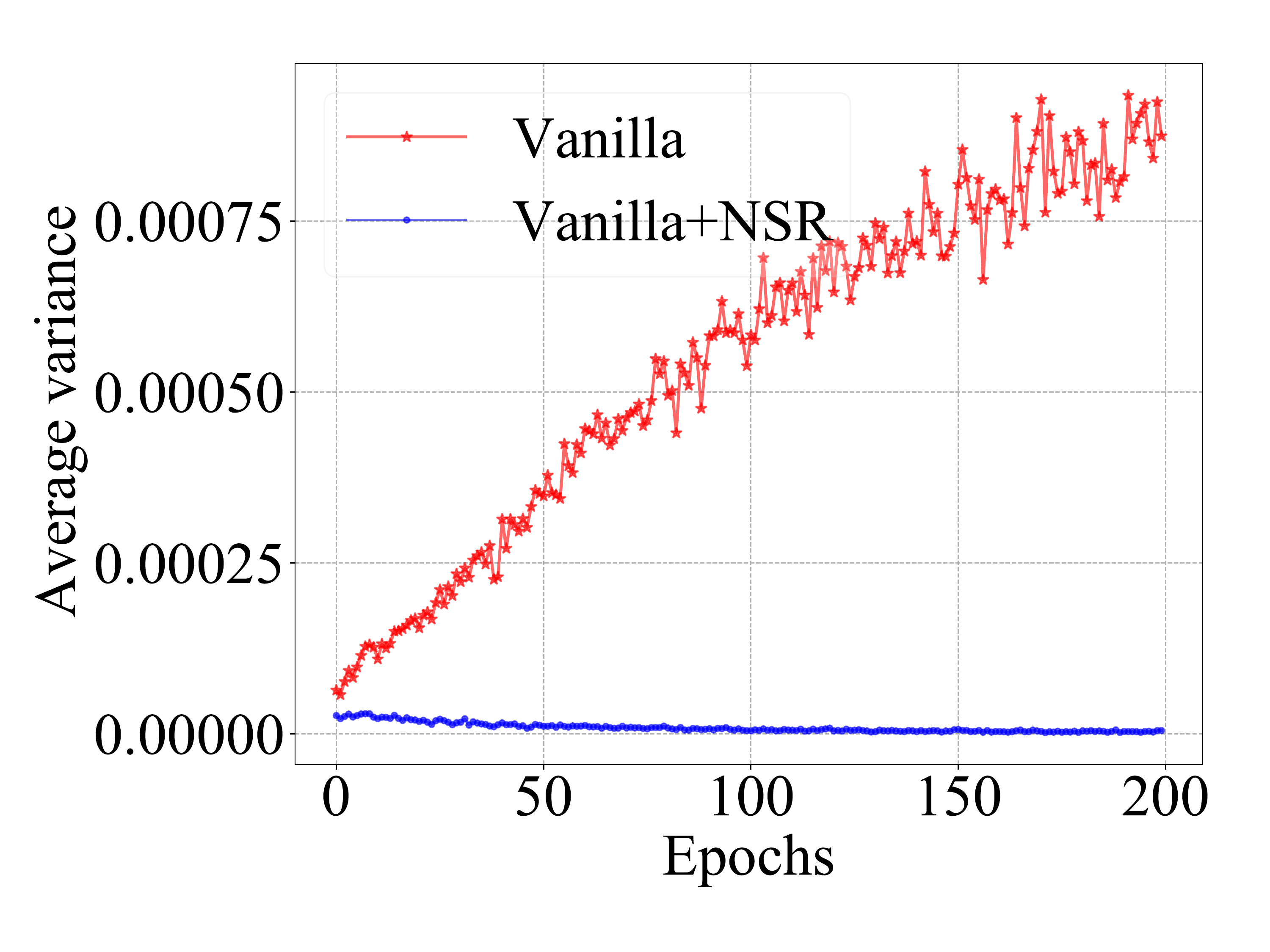}
        \end{minipage}
    }
    \caption{Training procedure of vanilla ResNet-18 and Resnet-18 with NSR on CIFAR-10. (a) and (b) illustrate the corresponding cross entropy loss and average intra-class response variance of these two models on training set.}
    \label{fig:hyperparam_CNN}
\end{figure*}

\begin{figure*}[!ht]
    \centering

    \subfigure[Cross entropy loss]{
        \begin{minipage}[b]{0.45\textwidth}
        \centering
        \includegraphics[width=1\textwidth]{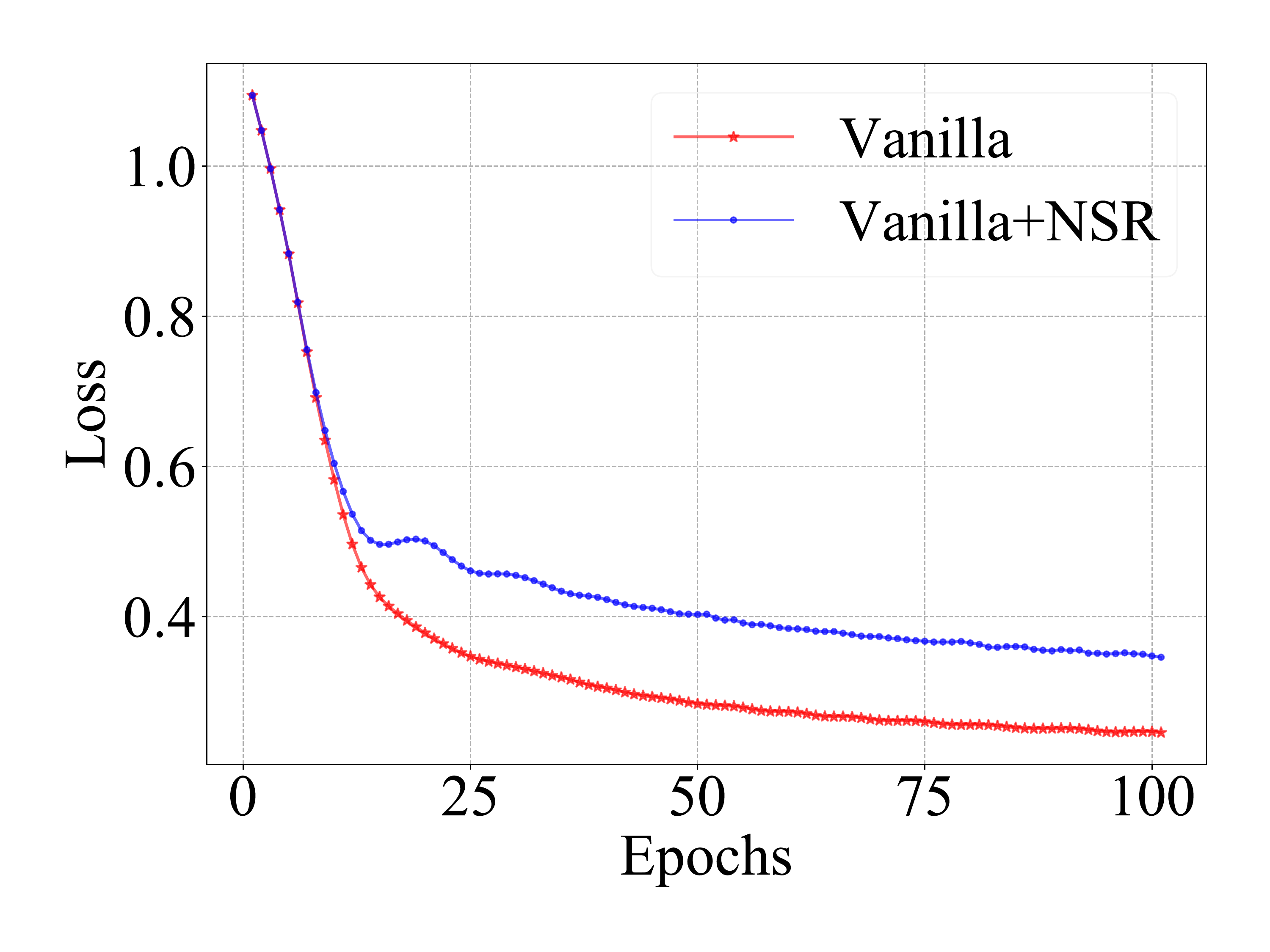}
        \end{minipage}
    }
    \subfigure[Intra-class variance]{
        \begin{minipage}[b]{0.45\textwidth}
        \centering
        \includegraphics[width=1\textwidth]{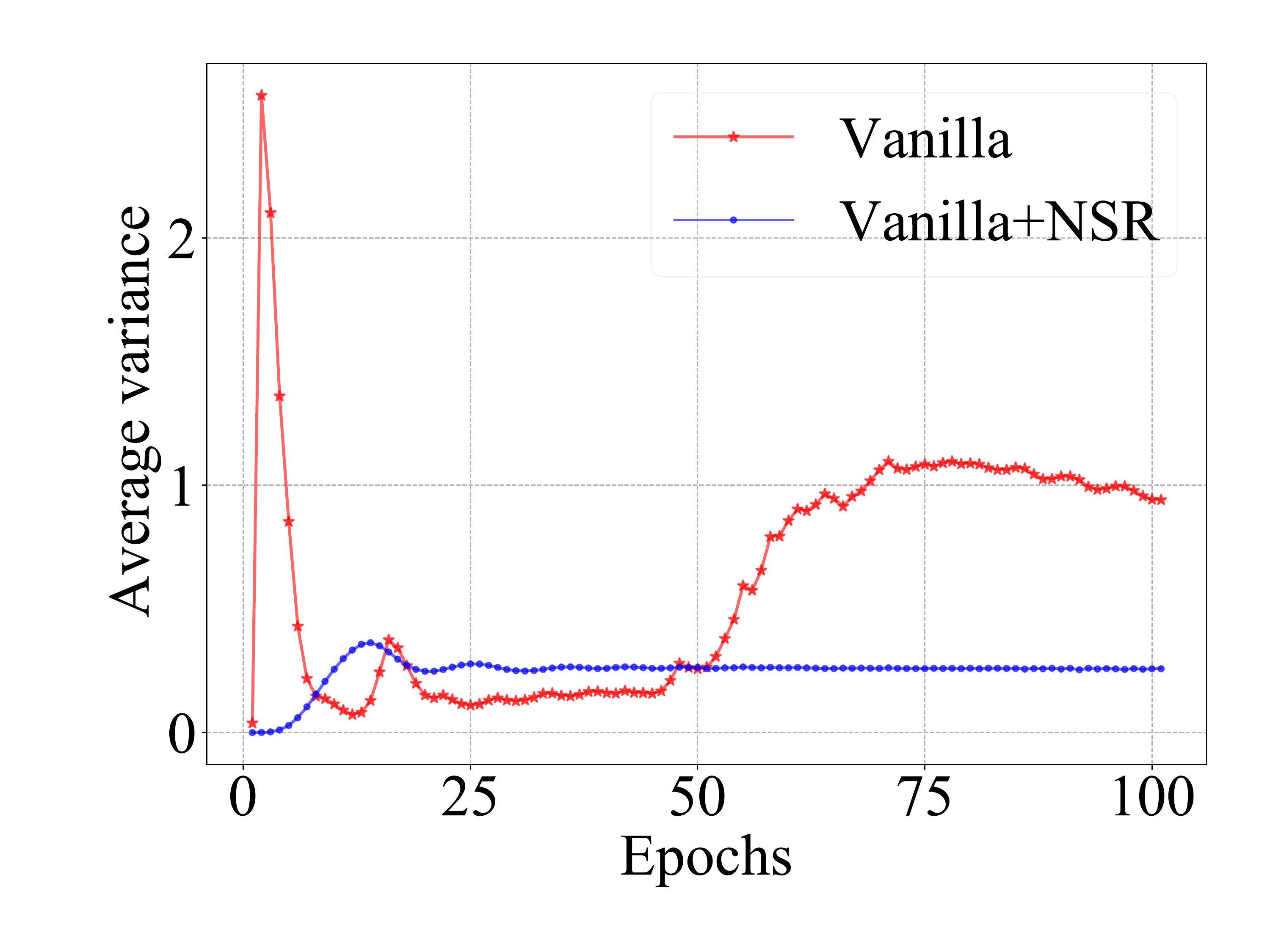}
        \end{minipage}
    }
    \caption{Training procedure of vanilla GraphSAGE-2 and GraphSAGE-2 with NSR on PubMed. (a) and (b) illustrate corresponding cross entropy loss and average intra-class response variance of these two models on the training set.}
    \label{fig:hyperparam_GNN}
\end{figure*}

\section{Dynamics of GNN and CNN during training procedure}\label{app:Dynamics}

In this section, we study the training procedure of ResNet-18 (CNN) and GraphSAGE-2 (GNN) of the vanilla version and model with NSR. The cross entropy loss and the expectation of neuron intra-class response variance  are demonstrated in Fig \ref{fig:hyperparam_CNN} and Fig \ref{fig:hyperparam_GNN}. The results of GraphSAGE-2 and ResNet-18 show a similar tendency with MLP-4 discussed in Section 2.2. Both figures illustrate that the model with NSR shows smaller intra-class response variance which leads to high test accuracy shown in Sec 4.2.1. 
The overall increasing tendency of intra-class variance on GraphSAGE-2 is similar to the tendency of MLP as expected. However, there exist a few outliers. A potential reason is that the sampling process of GraphSAGE will introduce noise in the generated features which leads to perturbed responses of individual neurons during the early training phase.

\section{Time and space complexity analysis \label{app:complexity}}

In this section, we elaborate on details of the time and space consumption of NSR.
The additional computation time overhead of adding NSR to the model is small. Tab. \ref{tab:time consumption} shows the comparison of time consumption for one training epoch between the vanilla models and models with NSR respectively on MLP, ResNet-18 (CNN) and GraphSAGE-2 (GNN).
We could see that the additional overhead is small especially for deeper networks like ResNet-18.
\begin{table}[!ht]
\centering
\caption{Time (s) consumption of vanilla models and models with NSR on one training epoch.}
\begin{tabular}{c|ccc}
\toprule
Model           & MLP-4           &ResNet-18 &GraphSAGE-2             \\ \midrule
Vanilla         & 2.19           &15.45    &0.0489                        \\ \midrule
Vanilla+NSR             & 2.54           &16.56    &0.0569                    \\ \bottomrule
\end{tabular}
\label{tab:time consumption}
\end{table}

\begin{figure}[ht]
    \centering
    \includegraphics[width=0.45\textwidth]{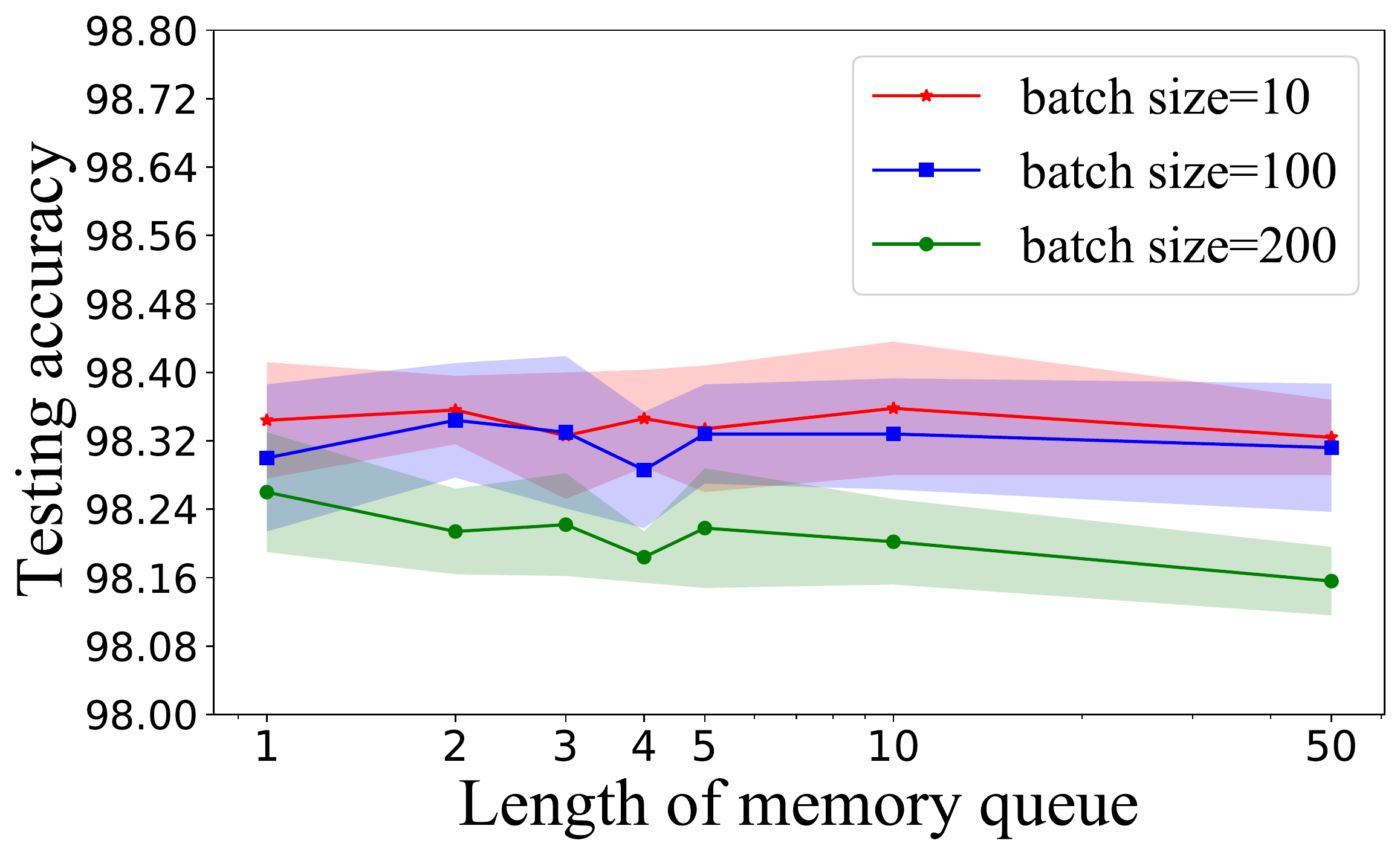}
    \caption{The testing accuracy of MLP-4 with NSR on MNIST with different memory queue length. The shadow indicates the standard deviation of the results obtained from five different runs with five different random seeds.}
    \label{fig:memory size}
\end{figure}
The additional space consumption is also small. The space complexity for NSR can be denoted as $O(N \times J \times M)$, where $N$ is the number of neurons in a layer, $J$ is the number of categories and $M$ is the length of memory queue. According to Fig. \ref{fig:memory size}, the performance of MLP model is robust with the length of memory queue and a small memory queue length can still lead to a competing performance. So the length of memory queue can be set as a small constant. All experiments set the length of memory queue as 5. Therefore, the computation space complexity can be directly denoted as $O(N \times J)$. We show that this space complexity is more negligible than that of vanilla model, including all model weights and intermediate outputs of every layer. Using MLP as an example, the space complexity of weights of a layer can be represented as $O(N_{in} \times N)$, where $N_{in}$ represents the input feature dimension. The space complexity of output for a layer can be denoted as $O(N \times B)$, where $B$ denotes batch size. Since the number of categories $J$ is usually smaller than the number of neurons $N_{in}$ and batch size $B$, the space complexity of NSR is smaller than that of weights and output for a layer.
The space used for NSR is usually much smaller than the space used for storing weights and layer output. 
\end{document}